\pdfoutput=1

\documentclass[11pt]{article}

\usepackage[]{acl}

\usepackage{times}
\usepackage{latexsym}

\usepackage[T1]{fontenc}

\usepackage[utf8]{inputenc}

\usepackage{microtype}

\usepackage{amsmath}
\usepackage{amssymb}
\usepackage{multirow}
\usepackage{multicol}
\usepackage{graphicx}
\usepackage{enumitem}
\usepackage{tikz}
\newcommand*{\circled}[1]{\lower.7ex\hbox{\tikz\draw (0pt, 0pt)%
    circle (.5em) node {\makebox[1em][c]{\scriptsize #1}};}}

\title{An Empirical Study of Consistency Regularization for End-to-End Speech-to-Text Translation}

\author{Pengzhi Gao, Ruiqing Zhang, Zhongjun He, Hua Wu, and Haifeng Wang \\
Baidu Inc. No. 10, Shangdi 10th Street, Beijing, 100085, China \\
\texttt{\{gaopengzhi,zhangruiqing01,hezhongjun\}@baidu.com} \\
\texttt{\{wu\_hua,wanghaifeng\}@baidu.com}
}

\begin{document}

\maketitle

\begin{abstract}
Consistency regularization methods, such as R-Drop \cite{NEURIPS2021_5a66b920} and CrossConST \cite{gao-etal-2023-improving}, have achieved impressive supervised and zero-shot performance in the neural machine translation (NMT) field. Can we also boost end-to-end (E2E) speech-to-text translation (ST) by leveraging consistency regularization? In this paper, we conduct empirical studies on intra-modal and cross-modal consistency and propose two training strategies, SimRegCR and SimZeroCR, for E2E ST in regular and zero-shot scenarios. Experiments on the MuST-C benchmark show that our approaches achieve state-of-the-art (SOTA) performance in most translation directions\footnote{Source code: https://github.com/gpengzhi/SimCR}. The analyses prove that regularization brought by the intra-modal consistency, instead of modality gap, is crucial for the regular E2E ST, and the cross-modal consistency could close the modality gap and boost the zero-shot E2E ST performance.
\end{abstract}

\section{Introduction}

Speech-to-text translation takes acoustic speech signals as input and outputs text translations in the target language. The conventional cascaded ST system consists of an automatic speech recognition (ASR) system and a machine translation (MT) module in a pipeline manner \cite{sperber-etal-2017-neural,sperber-etal-2019-self,zhang-etal-2019-lattice}. Recent works on ST have focused on the end-to-end system, which learns a unified model that directly generates text translations from speech without any intermediate outputs \cite{duong-etal-2016-attentional,berard2016listen}. E2E ST is a cross-modal task, where the major challenges include parallel ST data scarcity and representation discrepancy between speech and text modalities. In order to boost E2E ST training, the techniques utilized by existing approaches include pretraining \cite{wang-etal-2020-curriculum,xu-etal-2021-stacked}, multi-task learning \cite{ye21_interspeech,tang-etal-2021-improving}, knowledge distillation \cite{liu19d_interspeech,inaguma-etal-2021-source}, and cross-modal representation learning \cite{ye-etal-2022-cross,wang-etal-2022-discrete,fang-feng-2023-understanding}. However, most methods are far from being widely used due to the sophisticated model architecture, complicated algorithm implementation, and tedious hyperparameter search.

Consistency regularization has been widely adopted and shown great promise to improve NMT performance \cite{sato-etal-2019-effective,chen-etal-2021-manifold,NEURIPS2021_5a66b920,gao-etal-2022-bi,gao-etal-2023-improving}. Specifically, \citet{NEURIPS2021_5a66b920} introduce an intra-lingual consistency, R-Drop, to regularize dropout and improve the supervised NMT performance, and \citet{gao-etal-2023-improving} propose a cross-lingual consistency, CrossConST, to learn universal representations and boost the zero-shot NMT performance. Given the similar problem formulations between NMT and E2E ST, a natural question arises: \textit{Can we significantly improve E2E ST performance by leveraging simple consistency regularization?}

In this paper, our primary goal is to provide a simple, easy-to-reproduce, but tough-to-beat strategy for learning E2E ST models. Inspired by \citet{NEURIPS2021_5a66b920} and \citet{gao-etal-2023-improving}, we propose two strategies, SimRegCR and SimZeroCR, for training E2E ST models in regular and zero-shot scenarios. We show that intra-modal consistency is crucial for the regular setting, and cross-modal consistency is the key for closing the modality gap and boosting the zero-shot performance. The contributions of this paper can be summarized as follows:
\begin{itemize}
\item We conduct empirical studies on consistency regularization and propose two simple but effective strategies for learning E2E ST models in regular and zero-shot scenarios.

\item Experimental results show that our approaches achieve significant improvements on the MuST-C benchmark and outperform the current SOTA methods CRESS \cite{fang-feng-2023-understanding} and DCMA \cite{wang-etal-2022-discrete}.
\end{itemize}

\section{Background}

\subsection{End-to-End Speech-to-Text Translation}

Speech translation corpora usually contain speech-transcription-translation triples, which can be denoted as $\mathcal{S} = \{\mathbf{s}^i, \mathbf{x}^i, \mathbf{y}^i\}_{i=1}^{|\mathcal{S}|}$. $\mathbf{s}$ denotes the sequence of the audio wave, $\mathbf{x}$ is the transcription in the source language, and $\mathbf{y}$ represents the translation in the target language. $\mathcal{S}$ could be pairwise combined into three parallel corpora, $\mathcal{S}_{asr} = \{\mathbf{s}^i, \mathbf{x}^i\}_{i=1}^{|\mathcal{S}|}$, $\mathcal{S}_{mt} = \{\mathbf{x}^i, \mathbf{y}^i\}_{i=1}^{|\mathcal{S}|}$, and $\mathcal{S}_{st} = \{\mathbf{s}^i, \mathbf{y}^i\}_{i=1}^{|\mathcal{S}|}$, for ASR, MT, and ST tasks respectively. The goal of E2E ST is to generate translation $\mathbf{y}$ directly from the speech $\mathbf{s}$ without generating transcription $\mathbf{x}$, and the standard training objective is to minimize the empirical risk:
\begin{equation}\label{regular_st_ce}
\mathcal{L}^{st}_{ce}(\theta) =  \ell(f(\mathbf{s}, \mathbf{y}; \theta), \ddot{\mathbf{y}}),
\end{equation}
where $\ell$ denotes the cross-entropy loss, $\theta$ is a set of model parameters, $f(\mathbf{s}, \mathbf{y}; \theta)$ is a sequence of probability predictions, and $\ddot{\mathbf{y}}$ is a sequence of one-hot label vectors for $\mathbf{y}$. Directly modeling the speech-to-text mapping is nontrivial due to the representation discrepancy between speech and text modalities. To alleviate ST data sparsity, people usually include ASR and MT supervisions from $\mathcal{S}_{asr}$ and $\mathcal{S}_{mt}$ as well as external corpora for E2E ST task.

\subsection{Consistency Regularization for Neural Machine Translation}

\citet{NEURIPS2021_5a66b920} propose an intra-lingual consistency regularization, R-Drop, for boosting NMT performance by forcing the output distributions of different sub-models generated by dropout to be consistent with each other. For each sentence pair $(\mathbf{x}, \mathbf{y})$, the training objective is defined as:
\begin{equation}\label{r-drop}
\mathcal{L}_{R-Drop}(\theta) = \mathcal{L}^{mt}_{ce}(\theta) + \alpha \mathcal{L}^{mt}_{intra}(\theta),
\end{equation}
where
\begin{equation}\label{regular_mt_ce}
\mathcal{L}^{mt}_{ce}(\theta) = \ell(f(\mathbf{x}, \mathbf{y}; \theta), \ddot{\mathbf{y}}),
\end{equation}
\begin{equation}\label{regular_mt_intra}
\mathcal{L}^{mt}_{intra}(\theta) = \text{biKL}(f_1(\mathbf{x}, \mathbf{y}; \theta), f_2(\mathbf{x}, \mathbf{y}; \theta)),
\end{equation}
$f_1(\cdot)$ and $f_2(\cdot)$ denote the two forward passes of the same model $f(\cdot)$ with the dropout operation, $\text{biKL}(\cdot, \cdot)$ is the bidirectional Kullback-Leibler (KL) divergence of two distributions,
\begin{equation}
\text{biKL}(a, b) = (\text{KL}(a \| b) + \text{KL}(b \| a))/2,
\end{equation}
$\text{KL}(\cdot \| \cdot)$ denotes the KL divergence of two distributions, and $\alpha$ is a scalar hyper-parameter.

\citet{gao-etal-2023-improving} introduce a cross-lingual consistency regularization, CrossConST, for bridging the representation gap among different languages and improving zero-shot translation in multilingual NMT. For each sentence pair $( \mathbf{x}, \mathbf{y})$, the training objective is defined as:
\begin{equation}\label{crossconst}
\mathcal{L}_{CrossConST}(\theta) = \mathcal{L}^{mt}_{ce}(\theta) + \beta \mathcal{L}^{mt}_{cross}(\theta),
\end{equation}
where
\begin{equation}
\mathcal{L}^{mt}_{cross}(\theta) = \text{KL}(f(\mathbf{x}, \mathbf{y}; \theta) \| f(\mathbf{y}, \mathbf{y}; \theta)),
\end{equation}
and $\beta$ is a scalar hyper-parameter.

\section{Datasets and Baseline Settings}

\subsection{Dataset Description}

We initially consider \texttt{en}$\rightarrow$\texttt{de} translation for empirical study on consistency regularization in Section \ref{sec:methodology} and then show further experiments for other translation directions in Section \ref{sec:exps}. The detailed statistics of all datasets are summarized in Table \ref{dataset_statics}.

\subsubsection{ST Datasets}

We conduct experiments on MuST-C \cite{di-gangi-etal-2019-must}, which is a multilingual speech translation dataset containing audio recordings with the corresponding transcriptions and translations from English (\texttt{en}) to 8 languages: German (\texttt{de}), Spanish (\texttt{es}), French (\texttt{fr}), Italian (\texttt{it}), Dutch (\texttt{nl}), Portuguese (\texttt{pt}), Romanian (\texttt{ro}), and Russian (\texttt{ru}). 
We use \texttt{dev} and \texttt{tst-COMMON} as the validation and test sets respectively.

\subsubsection{MT Datasets}

We utilize external MT datasets to boost the E2E ST performance. Specifically, we incorporate WMT13 \cite{bojar-etal-2013-findings} dataset for \texttt{en}$\rightarrow$\texttt{es}, WMT14 \cite{bojar-etal-2014-findings} dataset for \texttt{en}$\rightarrow$\texttt{fr}, WMT16 \cite{bojar-etal-2016-findings} datasets for \texttt{en}$\rightarrow$\texttt{de}/\texttt{ro}/\texttt{ru}, and OPUS100 \cite{zhang-etal-2020-improving} datasets for \texttt{en}$\rightarrow$\texttt{it}/\texttt{nl}/\texttt{pt}. Note that we also use \texttt{dev} and \texttt{tst-COMMON} in the MuST-C dataset as the validation and test sets for the MT tasks. 

\begin{figure*}[h]
\centering
\includegraphics[scale=0.55]{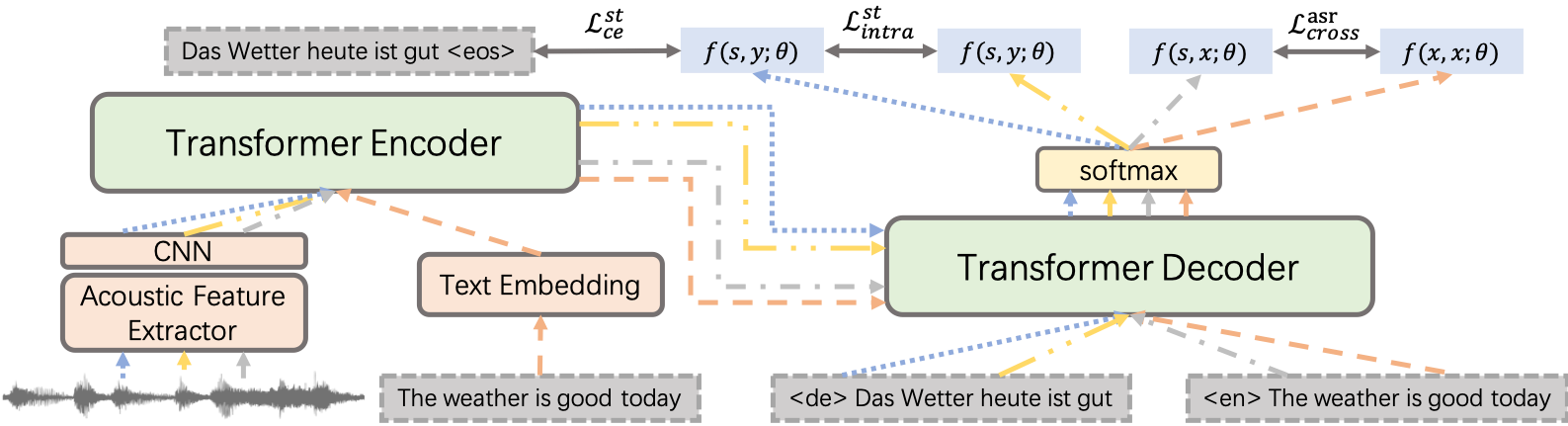}
\caption{Illustration of the intra-modal and cross-modal consistency regularization. For $\mathcal{L}^{st}_{intra}(\theta)$, the Speech-German pair (Speech, "Das Wetter heute ist gut") goes through the E2E ST model twice and obtain two output distributions $f(\mathbf{s}, \mathbf{y}; \theta)$. For $\mathcal{L}^{asr}_{cross}(\theta)$, the original Speech-English pair (Speech, "The weather is good today") and the copied English-English pair ("The weather is good today", "The weather is good today") go through the E2E ST model and the NMT model respectively and obtain two output distributions $f(\mathbf{s}, \mathbf{x}; \theta)$ and $f(\mathbf{x}, \mathbf{x}; \theta)$.}
\label{fig:architecture}
\end{figure*}


\subsection{Baseline Settings}

We adopt a widely used baseline model, W2V2-Transformer \cite{ye21_interspeech} in our empirical study (Figure \ref{fig:architecture}), which consists of a learnable acoustic feature extractor before two 1-dimensional convolutional layers and the standard Transformer architecture \cite{vaswani2017attention}. We use different language tags at the decoder input to distinguish the target languages. During inference, the language tag serves as the initial token to predict the output text. For example, if the speech input for the sentence ``The weather is good today'' is in English, to perform ASR, we use <\texttt{en}> as the initial token and decode ``The weather is good today'', while to translate into German, we use <\texttt{de}> as the initial token and decode ``Das Wetter heute ist gut''.

\paragraph{Pre-processing}

For speech input, we utilize the raw 16-bit 16kHz mono-channel audio wave. Following common practice, utterances with less than $1000$ frames are removed, and utterances with more than $480000$ frames are removed in the training set for GPU efficiency. For each translation direction, we jointly learn a unigram SentencePiece \cite{kudo-richardson-2018-sentencepiece} model with size $10$K on both the source and target sentences and use it to segment sentences into subwords for MT and ST tasks. For the external MT datasets, we filter out parallel sentences which length ratio exceeds $1.5$.

\paragraph{Model Configuration}

We use wav2vec2.0\footnote{\url{https://dl.fbaipublicfiles.com/fairseq/wav2vec/wav2vec\_small.pt}} \cite{NEURIPS2020_92d1e1eb} as the acoustic feature extractor, which is pretrained on the audio data from LibriSpeech \cite{panayotov2015librispeech}. Two 1-dimensional convolutional layers are added following the acoustic feature extractor, with kernel size 5, stride size 2, padding 2, and hidden dimension 1024. We utilize 6-layer transformer encoder and 6-layer transformer decoder. Each of the transformer layers comprises 512 hidden units, 8 attention heads, and 2048 feed-forward hidden units. 

\paragraph{Training Configuration}

We apply cross-entropy loss with label smoothing rate $0.1$ and set max tokens per batch to be $4096$ for the MT task and $2000000$ for the ASR and ST tasks. We use the Adam optimizer with Beta $(0.9, 0.98)$, $4000$, $8000$, and $4000$ warmup updates, and inverse square root learning rate scheduler with initial learning rate $1e^{-4}$, $1e^{-3}$, and $1e^{-4}$ for the ASR, MT, and ST tasks respectively. We apply the same configuration in each stage of the training procedure. 
During inference, we use beam search decoding with a beam size of $8$ with length penalty $1.2$, $0.6$, $1.8$, $1.0$, $1.0$, $1.4$, $1.4$, and $0.8$ for \texttt{en}$\rightarrow$\texttt{de}, \texttt{es}, \texttt{fr}, \texttt{it}, \texttt{nl}, \texttt{pt}, \texttt{ro}, and \texttt{ru}, respectively. We evaluate the MT and ST tasks by case-sensitive sacreBLEU \cite{post-2018-call}. We train all models until convergence on 8 NVIDIA Tesla V100 GPUs. 
For all the experiments below, we select the saved model state with the best validation performance.

\section{Methodology}\label{sec:methodology}

\begin{table*}\small
\centering
\begin{tabular}{c|c|c|c|c}
\hline
ID & Training Stage & Loss Function & MT BLEU & ST BLEU \\
\hline\hline
\circled{1} & MT train from scratch & $\mathcal{L}^{mt}_{ce}$ & 29.33 & - \\
\circled{2} & MT train from scratch & $\mathcal{L}^{mt}_{ce} + \alpha\mathcal{L}^{mt}_{intra}$ & 32.76 & - \\
\hline
\circled{3} & ST train from scratch & $\mathcal{L}^{st}_{ce}$ & - & 23.49 \\
\circled{4} & ST train from scratch & $\mathcal{L}^{st}_{ce} + \alpha\mathcal{L}^{st}_{intra}$ & - & 26.77 \\
\circled{5} & ST finetune on \circled{1} & $\mathcal{L}^{st}_{ce}$ & - & 24.38 \\
\circled{6} & ST finetune on \circled{1} & $\mathcal{L}^{st}_{ce} + \alpha\mathcal{L}^{st}_{intra}$ & - & 27.35 \\
\circled{7} & ST finetune on \circled{2} & $\mathcal{L}^{st}_{ce} + \alpha\mathcal{L}^{st}_{intra}$ & - & \bf 27.91 \\
\hline
\circled{8} & MT \& ST train from scratch & $\mathcal{L}^{mt}_{ce} + \mathcal{L}^{st}_{ce}$ & 28.54 & 23.75 \\
\circled{9} & MT \& ST finetune on \circled{1} & $\mathcal{L}^{mt}_{ce} + \mathcal{L}^{st}_{ce}$ & 29.73 & 23.82 \\
\circled{10} & MT \& ST finetune on \circled{1} & $\mathcal{L}^{mt}_{ce} + \mathcal{L}^{st}_{ce} + \beta\mathcal{L}^{mt-st}_{cross}$ & 30.66 & 26.87 \\
\circled{11} & MT \& ST finetune on \circled{2} & $\mathcal{L}^{mt}_{ce} + \alpha\mathcal{L}^{mt}_{intra} + \mathcal{L}^{st}_{ce} + \alpha\mathcal{L}^{st}_{intra}$ & 32.70 & 27.48 \\
\circled{12} & MT \& ST finetune on \circled{11} & $\mathcal{L}^{mt}_{ce} + \alpha\mathcal{L}^{mt}_{intra} + \mathcal{L}^{st}_{ce} + \alpha\mathcal{L}^{st}_{intra} + \beta\mathcal{L}^{mt-st}_{cross}$ & 31.00 & 27.57 \\
\hline
\circled{13} & MT train from scratch$^{\dagger}$ & $\mathcal{L}^{mt}_{ce}$ & 29.61 & - \\
\circled{14} & MT train from scratch$^{\dagger}$ & $\mathcal{L}^{mt}_{ce} + \alpha\mathcal{L}^{mt}_{intra}$ & 30.02 & - \\
\circled{15} & MT finetune on \circled{13} & $\mathcal{L}^{mt}_{ce}$ & 33.59 & - \\
\circled{16} & MT finetune on \circled{14} & $\mathcal{L}^{mt}_{ce} + \alpha\mathcal{L}^{mt}_{intra}$ & 34.11 & - \\
\hline
\circled{17} & ST finetune on \circled{15} & $\mathcal{L}^{st}_{ce}$ & - & 27.33 \\
\circled{18} & ST finetune on \circled{15} & $\mathcal{L}^{st}_{ce} + \alpha\mathcal{L}^{st}_{intra}$ & - & 28.96 \\
\circled{19} & ST finetune on \circled{16} & $\mathcal{L}^{st}_{ce} + \alpha\mathcal{L}^{st}_{intra}$ & - & \bf 29.23 \\
\end{tabular}
\caption{Case-sensitive detokenized BLEU scores on the MuST-C \texttt{en}$\rightarrow$\texttt{de} \texttt{tst-COMMON} set. $\dagger$ denotes the MT training is performed on the WMT16 dataset, other MT training is performed on the MuST-C dataset. We mark the best ST BLEU scores in two experimental setups in bold. The choices for $\alpha$ and $\beta$ are summarized in Table \ref{tab:es_regular_st_hyperparameter}.}\label{tab:es_regular_st}
\end{table*}

In this section, we formally propose SimRegCR and SimZeroCR, the consistency-based strategies for learning E2E ST models in regular (Section \ref{sec:regular_e2e_st}) and zero-shot (Section \ref{sec:zeroshot_e2e_st}) scenarios respectively. We introduce the details of each part below.

\subsection{Consistency Regularization for Regular End-to-End Speech Translation}\label{sec:regular_e2e_st}

We here investigate the performance of consistency regularization for the regular scenario, where we learn the E2E ST model by utilizing MT and ST datasets. For each training sample, the loss functions include: $\mathcal{L}^{mt}_{ce}(\theta)$, $\mathcal{L}^{mt}_{intra}(\theta)$, $\mathcal{L}^{st}_{ce}(\theta)$,
\begin{equation}\label{regular_st_intra}
\mathcal{L}^{st}_{intra}(\theta) = \text{biKL}(f_1(\mathbf{s}, \mathbf{y}; \theta), f_2(\mathbf{s}, \mathbf{y}; \theta)),
\end{equation}
and
\begin{equation}\label{regular_mt_st_cross}
\mathcal{L}^{mt-st}_{cross}(\theta) = \text{KL}(f(\mathbf{x}, \mathbf{y}; \theta) \| f(\mathbf{s}, \mathbf{y}; \theta)),
\end{equation}
where \eqref{regular_st_ce} and \eqref{regular_mt_ce} are the cross-entropy loss for the ST and MT tasks respectively, \eqref{regular_mt_intra} and \eqref{regular_st_intra} are the intra-modal consistency regularization for the MT and ST tasks respectively, and \eqref{regular_mt_st_cross} denotes the cross-modal consistency regularization between the MT and ST tasks, which could also be regarded as the sequence-level knowledge distillation from the MT model to the ST model \cite{liu19d_interspeech}.

\subsubsection{Experimental Results}

We consider two experimental setups: without external MT data (\circled{1} - \circled{12}) and with external MT data (\circled{13} - \circled{19}), and summarize the experimental results in Table \ref{tab:es_regular_st}. Note that \circled{5} and \circled{17} correspond to the W2V2-Transformer baselines in the settings of without and with external MT data respectively. By checking model performance under different combinations of loss function and training strategy, we have the following observations: 1) The intra-modal consistency, $\mathcal{L}^{mt}_{intra}$ and $\mathcal{L}^{st}_{intra}$, could boost the MT (\circled{1} vs \circled{2}; \circled{13} vs \circled{14}) and ST (\circled{3} vs \circled{4}) performance. 2) The paradigm of pretraining-finetuning could further improve the ST performance (\circled{3} vs \circled{5}; \circled{4} vs \circled{7}). 3) The multi-task learning achieves similar performance compared with the pretraining-finetuning strategy (\circled{3} vs \circled{8}; \circled{5} vs \circled{9}). 4) The cross-modal consistency, $\mathcal{L}^{mt-st}_{cross}$, could improve the ST performance (\circled{9} vs \circled{10}; \circled{11} vs \circled{12}) but still achieve the sub-optimal performance (\circled{7} vs \circled{12}).

\subsubsection{Does Intra-modal Consistency Implicitly Bridge the Modality Gap?}

\begin{figure}[h]
\centering
\includegraphics[scale=0.33]{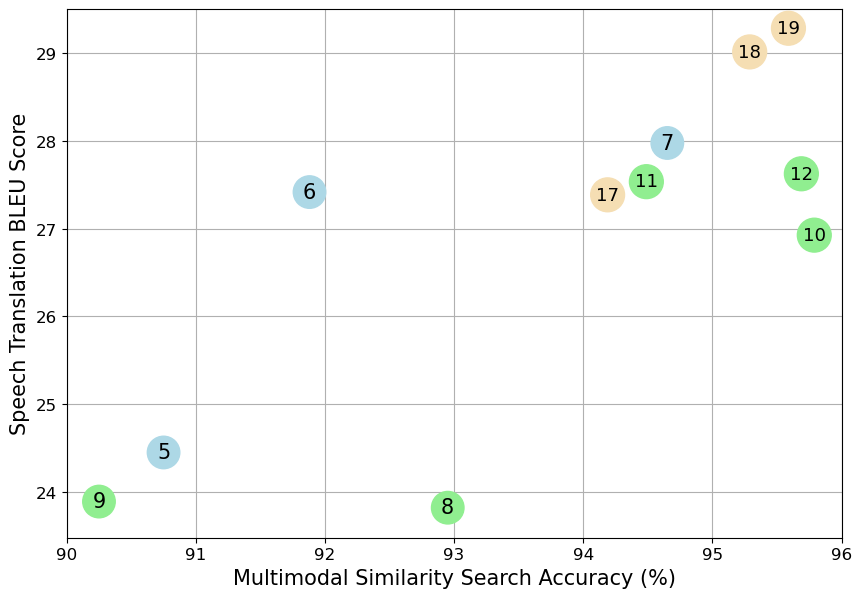}
\caption{The ST BLEU score and similarity search accuracy of each model in Table \ref{tab:es_regular_st} on the MuST-C \texttt{en}$\rightarrow$\texttt{de} \texttt{tst-COMMON} set. The blue circles denote the pretraining-finetuning experiments without external MT data. The green circles denote the multi-task learning experiments without external MT data. The orange circles denote the experiments with external MT data.}
\label{fig:multimodal_ss}
\end{figure}

One interesting finding from the empirical study is that the strategies (\circled{7} and \circled{19}) only utilizing the intra-modal consistency achieve the best ST performance instead of explicitly leveraging the cross-modal consistency. We here investigate the impact of the consistency regularization on the modality gap and the E2E ST performance. We conduct a multimodal similarity search experiment and use the averaged bidirectional similarity search accuracy as the metric to evaluate the modality gap. Given parallel speech-transcription pairs, we find the nearest neighbor for each one in the other modality according to the representation cosine similarity and compute the corresponding accuracy, where the speech and transcription representations are calculated by max-pooling the encoder outputs. The evaluation results are reported in Figure \ref{fig:multimodal_ss}. By checking the relationship between ST BLEU score and multimodal similarity search accuracy, we have the following observations: 1) The intra-modal consistency, $\mathcal{L}^{mt}_{intra}$ and $\mathcal{L}^{st}_{intra}$, implicitly closes the modality gap (\circled{5} vs \circled{6} vs \circled{7}; \circled{17} vs \circled{18} vs \circled{19}). 2) The cross-modal consistency, $\mathcal{L}^{mt-st}_{cross}$, explicitly bridges the modality gap (\circled{9} vs \circled{10}; \circled{11} vs \circled{12}). 3) A closer modality gap does not guarantee a better ST performance (\circled{6} vs \circled{10}; \circled{7} vs \circled{12}), and the regularization effect introduced by the intra-modal consistency seems to be more crucial for the regular E2E ST task which is in line with \citet{han-etal-2023-modality}.

\subsubsection{Training Strategy}

\begin{figure}[h]
\centering
\includegraphics[scale=0.45]{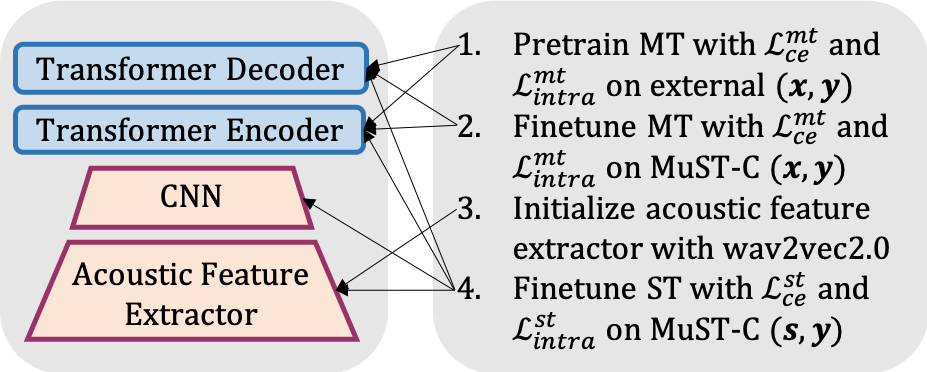}
\caption{The training steps of SimRegCR by utilizing the intra-modal consistency regularization. In each step, the modules that contribute to the final E2E ST model are pointed out by arrow lines. We also consider SimRegCR$^-$ (\circled{18} in Table \ref{tab:es_regular_st}) in this paper, which trains MT model only with $\mathcal{L}^{mt}_{ce}$ in the first two steps.}
\label{fig:regcr_ts}
\end{figure}

We here summarize the multi-stage training strategy, SimRegCR (\circled{19} in Table \ref{tab:es_regular_st}), consisting of MT pretraining and ST finetuning with the intra-modal consistency regularization in Figure \ref{fig:regcr_ts}. The setting without external MT data only differs by removing the first step of external MT pretraining. 

\paragraph{Comparison with Existing Methods}

\begin{table}\small
\centering
\begin{tabular}{l|c c}
\hline
Method & \multicolumn{2}{c}{BLEU} \\
 & w/o WMT16 & w/ WMT16 \\
\hline\hline
XSTNet$^{\dagger}$ & 25.2 & 27.1 \\
STEMM$^{\dagger}$ & 25.6 & 28.7 \\
ConST$^{\dagger}$ & 25.7 & 28.3 \\
CMOT$^{\dagger}$ & 27.0 & 29.0 / 28.5$^*$ \\
CRESS$^{\dagger}$ & 27.2 & {\bf 29.4} / 28.9$^*$ \\
\hline
W2V2-Transformer & 24.4 & 27.3 \\
\ \ + SimRegCR$^-$ & 27.4 & 29.0 \\
\ \ + SimRegCR & \bf 27.9 & 29.2 \\
\end{tabular}
\caption{Our method achieves the superior or comparable performance over the existing methods on the MuST-C \texttt{en}$\rightarrow$\texttt{de} benchmark. $*$ denotes the performance using wav2vec2.0 instead of HuBERT as the acoustic feature extractor. $\dagger$ denotes the numbers are reported from the corresponding papers, others are based on our runs.\label{regular_st_benchmark}}
\end{table}

We summarize the recent results of several existing works on the MuST-C \texttt{en}$\rightarrow$\texttt{de} benchmark in Table \ref{regular_st_benchmark}. The existing methods vary from different aspects, including cross-modal progressive training (XSTNet) \cite{ye21_interspeech}, cross-modal manifold mixup (STEMM) \cite{fang-etal-2022-stemm}, cross-modal contrastive learning (ConST) \cite{ye-etal-2022-cross}, cross-modal mixup via optimal transport (CMOT) \cite{zhou-etal-2023-cmot}, and cross-modal regularization with scheduled sampling (CRESS) \cite{fang-feng-2023-understanding}. Note that XSTNet, STEMM, and ConST adopt wav2vec2.0 as the acoustic feature extractor, while CMOT and CRESS use HuBERT \cite{10.1109/TASLP.2021.3122291} which could achieve slightly stronger baseline. We can see that SimRegCR$^-$ achieves an improvement of $2.35$ BLEU score on average over W2V2-Transformer, and SimRegCR achieves the superior or comparable performance over the current SOTA method CRESS that incorporates cross-modal regularization, scheduled sampling, token-level adaptive training, and a stronger acoustic feature extractor. 

\subsection{Consistency Regularization for Zero-shot End-to-End Speech Translation}\label{sec:zeroshot_e2e_st}

We here investigate the performance of consistency regularization for the zero-shot scenario, where we learn the E2E ST model by utilizing ASR and MT datasets. For each training sample, the loss functions include: $\mathcal{L}^{mt}_{ce}(\theta)$, $\mathcal{L}^{mt}_{intra}(\theta)$,
\begin{equation}\label{zs_asr_ce}
\mathcal{L}^{asr}_{ce}(\theta) = \ell(f(\mathbf{s}, \mathbf{x}; \theta), \ddot{\mathbf{x}}),
\end{equation}
\begin{equation}\label{zs_asr_intra}
\mathcal{L}^{asr}_{intra}(\theta) = \text{biKL}(f_1(\mathbf{s}, \mathbf{x}; \theta), f_2(\mathbf{s}, \mathbf{x}; \theta)),
\end{equation}
and
\begin{equation}\label{zs_asr_cross}
\mathcal{L}^{asr}_{cross}(\theta) = \text{KL}(f(\mathbf{s}, \mathbf{x}; \theta) \| f(\mathbf{x}, \mathbf{x}; \theta)),
\end{equation}
where \eqref{regular_mt_ce} and \eqref{zs_asr_ce} are the cross-entropy loss for the MT and ASR tasks respectively, \eqref{regular_mt_intra} and \eqref{zs_asr_intra} are the intra-modal consistency regularization for the MT and ASR tasks respectively, and \eqref{zs_asr_cross} denotes the cross-modal consistency regularization for the ASR task, which could be regarded as the multimodal version of CrossConST \cite{gao-etal-2023-improving}.

\begin{table*}\small
\centering
\begin{tabular}{c|c|c|c|c}
\hline
ID & Training Stage & Loss Function & MT BLEU & ST BLEU \\
\hline\hline
\circled{1} & MT train from scratch$^{\dagger}$ & $\mathcal{L}^{mt}_{ce}$ & 29.61 & - \\
\circled{2} & MT train from scratch$^{\dagger}$ & $\mathcal{L}^{mt}_{ce} + \alpha\mathcal{L}^{mt}_{intra}$ & 30.02 & - \\
\hline
\circled{3} & MT Finetune on \circled{1} & $\mathcal{L}^{mt}_{ce}$ & 33.59 & - \\
\circled{4} & MT Finetune on \circled{2} & $\mathcal{L}^{mt}_{ce} + \alpha\mathcal{L}^{mt}_{intra}$ & 34.11 & - \\
\hline
\circled{5} & ASR \& MT finetune on \circled{3} & $\mathcal{L}^{asr}_{ce} + \mathcal{L}^{mt}_{ce}$ & 33.99 & 0.46 \\
\circled{6} & ASR \& MT finetune on \circled{3} & $\mathcal{L}^{asr}_{ce} + \mathcal{L}^{mt}_{ce} + \beta\mathcal{L}^{asr}_{cross}$ & 32.82 & \bf 25.10 \\
\hline
\circled{7} & ASR \& MT finetune on \circled{4} & $\mathcal{L}^{asr}_{ce} + \alpha\mathcal{L}^{asr}_{intra} + \mathcal{L}^{mt}_{ce} + \alpha\mathcal{L}^{mt}_{intra}$ & 34.35 & 0.56 \\
\circled{8} & ASR \& MT finetune on \circled{7} & $\mathcal{L}^{asr}_{ce} + \alpha\mathcal{L}^{asr}_{intra} + \mathcal{L}^{mt}_{ce} + \alpha\mathcal{L}^{mt}_{intra} + \beta\mathcal{L}^{asr}_{cross}$ & 33.25 & 24.86 \\
\end{tabular}
\caption{Case-sensitive detokenized BLEU scores on the MuST-C \texttt{en}$\rightarrow$\texttt{de} \texttt{tst-COMMON} set. $\dagger$ denotes the MT training is performed on the WMT16 dataset, other MT training is performed on the MuST-C dataset. We mark the best ST BLEU score in bold. The choices for $\alpha$ and $\beta$ are summarized in Table \ref{tab:es_zeroshot_st_hyperparameter}.}\label{tab:es_zeroshot_st}
\end{table*}

\subsubsection{Experimental Results}

We consider the experimental setup with external MT data and summarize the experimental results in Table \ref{tab:es_zeroshot_st}. Note that \circled{5} corresponds to the W2V2-Transformer baseline. By checking model performance under different combinations of loss function and training strategy, we have the following observations: 1) The cross-modal consistency, $\mathcal{L}^{asr}_{cross}$, could boost the zero-shot ST performance (\circled{5} vs \circled{6}; \circled{7} vs \circled{8}). 2) Leveraging the intra-modal consistency, $\mathcal{L}^{asr}_{intra}$ and $\mathcal{L}^{mt}_{intra}$, could improve the corresponding MT performance (\circled{5} vs \circled{7}; \circled{6} vs \circled{8}), but could not achieve the superior performance in the zero-shot ST direction (\circled{6} vs \circled{8}).

\subsubsection{Does the Cross-modal Consistency Really Close the Modality Gap?}

\begin{figure}[h]
\centering
\includegraphics[scale=0.36]{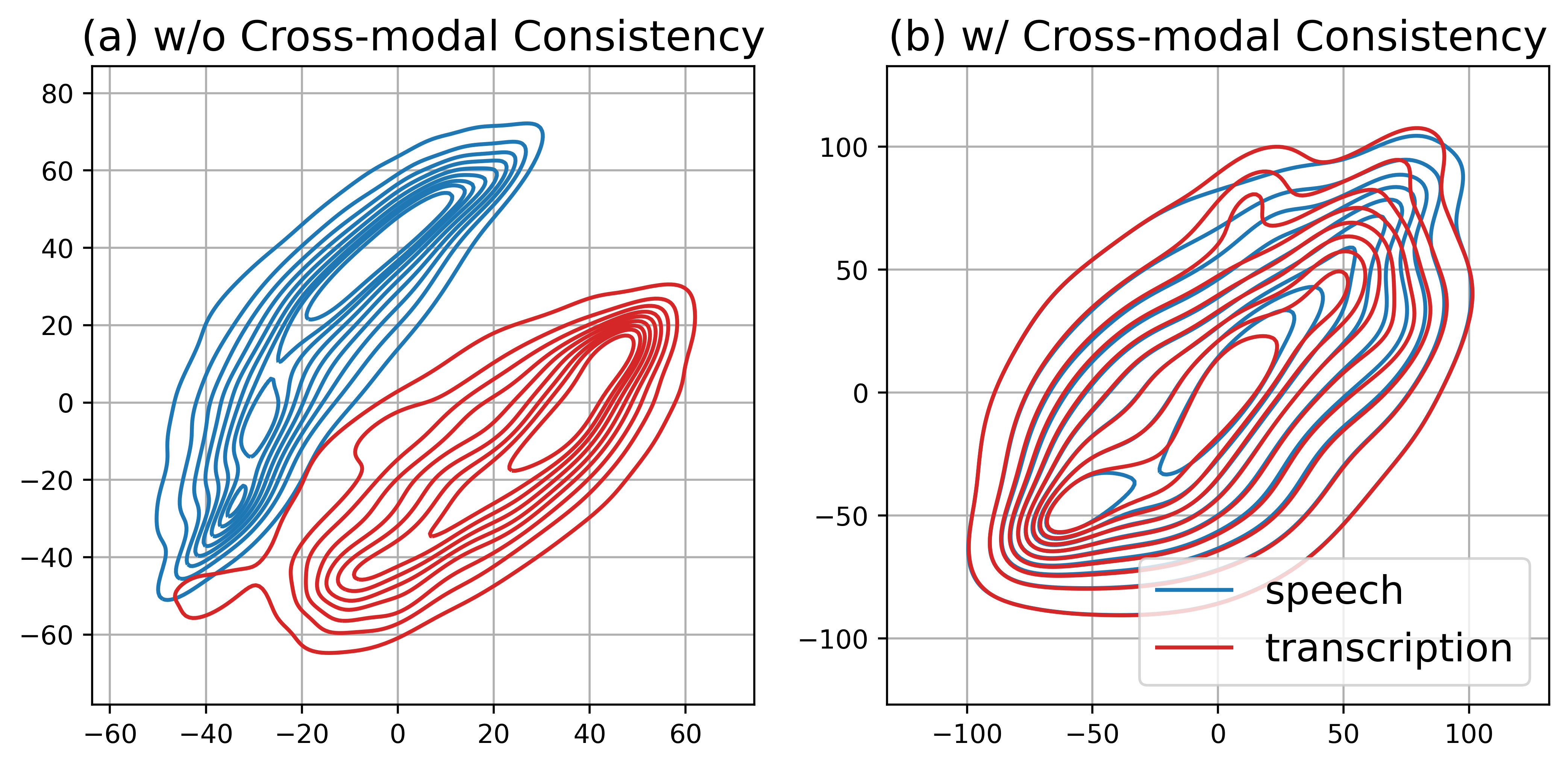}
\caption{Bivariate kernel density estimation plots of the speech and transcription representations after using T-SNE dimensionality reduction, where the max-pooled outputs of the W2V2-Transformer encoder are applied as the speech and transcription representations. }
\label{fig:multimodal_bkd}
\end{figure}

To verify whether the cross-modal consistency regularization can better align the modality representation space, we visualize the speech and transcription representations of the MuST-C \texttt{en}$\rightarrow$\texttt{de} \texttt{tst-COMMON} set. We apply dimension reduction on the 512-dimensional representations with T-SNE \cite{NIPS2002_6150ccc6} and then depict the bivariate kernel density estimation based on the 2-dimensional representations in Figure \ref{fig:multimodal_bkd}. Figure \ref{fig:multimodal_bkd} shows that the W2V2-Transformer baseline (\circled{5}) cannot align speech and transcription well in the representation space, while the cross-modal consistency (\circled{6}) draws the representations across different modalities much closer.

\subsubsection{Training Strategy}

\begin{figure}[h]
\centering
\includegraphics[scale=0.45]{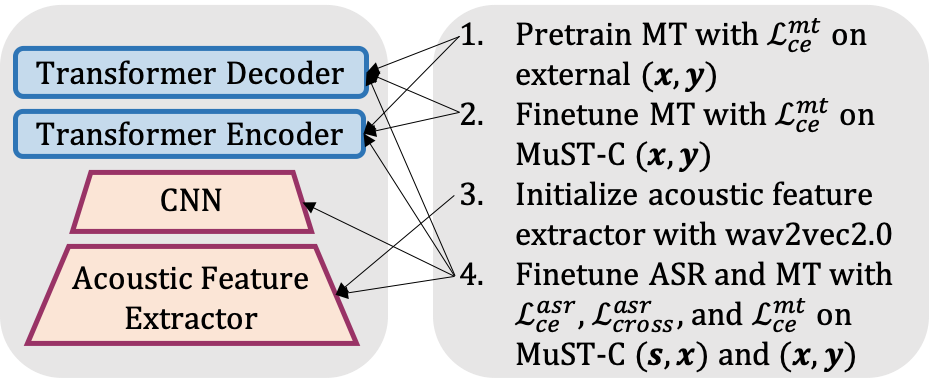}
\caption{The training steps of SimZeroCR by utilizing the cross-modal consistency regularization. In each step, the modules that contribute to the final E2E ST model are pointed out by arrow lines.}
\label{fig:zerocr_ts}
\end{figure}

We here summarize the multi-stage training strategy, SimZeroCR (\circled{6} in Table \ref{tab:es_zeroshot_st}), consisting of MT pretraining and ASR \& MT finetuning with the cross-modal consistency regularization in Figure \ref{fig:zerocr_ts}.

\begin{table}\small
\centering
\begin{tabular}{l|c c c|c}
\hline
Method & \multicolumn{3}{c|}{Training Data} & BLEU \\
 & Speech & ASR & MT & \\
\hline\hline
MultiSLT$^{\dagger}$ & - & \checkmark & \checkmark & 6.8 \\
Chimera$^{\dagger}$ & \checkmark & \checkmark & \checkmark & 13.5 \\
DCMA$^{\dagger}$ & \checkmark & \checkmark & \checkmark & 24.0 \\
\hline
W2V2-Transformer & \checkmark & \checkmark & \checkmark & 0.5 \\
\ \ + SimZeroCR & \checkmark & \checkmark & \checkmark & \bf 25.1 \\
\end{tabular}
\caption{Our method achieves the superior performance over the existing methods on the MuST-C \texttt{en}$\rightarrow$\texttt{de} benchmark. $\dagger$ denotes the numbers are reported from \citet{wang-etal-2022-discrete}, others are based on our runs.}\label{zeroshot_st_benchmark}
\end{table}

\paragraph{Comparison with Existing Methods}

\begin{table*}[h]\small
\centering
\begin{tabular}{l | c | c c c c c c c c} 
Method & External & \multicolumn{8}{|c}{BLEU} \\
 & Speech & \texttt{de} & \texttt{es} & \texttt{fr} & \texttt{it} & \texttt{nl} & \texttt{pt} & \texttt{ro} & \texttt{ru} \\
\hline
\hline
Fairseq ST \cite{wang-etal-2020-fairseq} & - & 22.7 & 27.2 & 32.9 & 22.7 & 27.3 & 28.1 & 21.9 & 15.3 \\
Dual Decoder \cite{le-etal-2020-dual} & - & 23.6 & 28.1 & 33.5 & 24.2 & 27.6 & 30.0 & 22.9 & 15.2 \\
Speechformer \cite{papi-etal-2021-speechformer} & - & 23.6 & 28.5 & - & - & 27.7 & - & - & - \\
SATE \cite{xu-etal-2021-stacked} & - & 25.2 & - & - & - & - & - & - & - \\
BiKD \cite{inaguma-etal-2021-source} & - & 25.3 & - & 35.3 & - & - & - & - & - \\ 
XSTNet \cite{ye21_interspeech} & \checkmark & 25.5 & 29.6 & 36.0 & 25.5 & 30.0 & 31.3 & 25.1 & 16.9 \\
STEMM \cite{fang-etal-2022-stemm} & \checkmark & 25.6 & 30.3 & 36.1 & 25.6 & 30.1 & 31.0 & 24.3 & 17.1 \\
ConST \cite{ye-etal-2022-cross} & \checkmark & 25.7 & 30.4 & 36.8 & 26.3 & 30.6 & 32.0 & 24.8 & 17.3 \\
FCCL$^{m}$ \cite{10042965} & \checkmark & 25.9 & 30.7 & 36.8 & 26.4 & 30.5 & 31.8 & 25.0 & 17.6 \\
M$^3$ST \cite{10095090} & \checkmark & 26.4 & 31.0 & 37.2 & 26.6 & 30.9 & 32.8 & 25.4 & 18.3 \\
CMOT \cite{zhou-etal-2023-cmot} & \checkmark & 27.0 & 31.1 & 37.3 & 26.9 & 31.2 & 32.7 & 25.3 & 17.9 \\
CRESS \cite{fang-feng-2023-understanding} & \checkmark & 27.2 & 31.9 & 37.8 & 27.3 & 31.6 & 33.0 & 25.9 & 18.7 \\
\hline
W2V2-Transformer & \checkmark & 24.4 & 29.9 & 34.7 & 25.1 & 29.3 & 30.3 & 23.4 & 16.5 \\
\ \ + SimRegCR$^-$ & \checkmark & 27.4 & 31.5 & 38.1 & 27.2 & 32.0 & 33.3 & 25.9 & 18.8 \\
\ \ + SimRegCR & \checkmark & $\bf 27.9^*$ & $\bf 32.1^*$ & $\bf 39.0^*$ & $\bf 27.7^*$ & $\bf 32.4^*$ & $\bf 34.0^*$ & $\bf 26.3^*$ & $\bf 19.0^*$ \\
\end{tabular}
\caption{Case-sensitive detokenized BLEU scores on MuST-C \texttt{tst-COMMON} set without external MT datasets. "External speech" denotes unlabeled speech data. * indicates the improvements over W2V2-Transformer are statistically significant with $p < 0.01$. The highest BLEU scores are marked in bold for all methods in each column.}\label{tab:res_without_mt}
\end{table*}

We summarize the recent results of several existing works on MuST-C \texttt{en}$\rightarrow$\texttt{de} benchmark in Table \ref{zeroshot_st_benchmark}. The existing methods vary from different aspects, including language-specific encoders-decoders architecture (MultiSLT) \cite{9688026}, continuous cross-modal alignment (Chimera) \cite{han-etal-2021-learning}, and discrete cross-modal alignment (DCMA) \cite{wang-etal-2022-discrete}. SimZeroCR achieves an improvement of $24.6$ BLEU score over W2V2-Transformer and outperforms the current SOTA method DCMA\footnote{Note that the external MT dataset and the inference configurations used in this section are slightly different from those used in \citet{wang-etal-2022-discrete}. Please check the experimental results in Section \ref{sec:zs_e2e_st} for more fair comparison.} that incorporates shared memory and vector quantization modules. 

\section{Experiments on More Languages}\label{sec:exps}


\subsection{Regular End-to-End Speech Translation}

We consider two experimental setups: without external MT data and with external MT data. The detailed information of the baseline methods are summarized in Appendix \ref{sec:reg_e2e_st_baseline}, and the BLEU scores of the baseline methods are reported from the corresponding papers. The choice for hyperparameters and the corresponding model performance in each training step of our approaches are summarized in Tables \ref{tab:11}, \ref{tab:12}, \ref{tab:13}, and \ref{tab:14}.

When there is no external MT data (Table \ref{tab:res_without_mt}), SimRegCR$^-$ gains an average improvement of $2.6$ BLEU scores over the W2V2-Transformer baseline and can achieve comparable performance to the current SOTA method CRESS. It is also worth mentioning that SimRegCR gains an average improvement of $3.1$ BLEU scores over the W2V2-Transformer baseline and achieves an average improvement of $0.6$ BLEU scores over CRESS that incorporates cross-modal regularization, scheduled sampling, token-level adaptive training, and a stronger acoustic feature extractor, which clearly shows the effectiveness of our methods. When external MT data is included (Table \ref{tab:res_with_mt}), SimRegCR$^-$ and SimRegCR gain average improvement of $1.7$ and $2.2$ BLEU scores over the W2V2-Transformer baseline respectively, and SimRegCR achieves an average improvement of $0.2$ BLEU scores over CRESS, which implies that we could easily achieve SOTA performance for E2E ST task by leveraging simple intra-modal consistency regularization.

\begin{table*}[h]\small
\centering
\begin{tabular}{l | c | c c c c c c c c} 
Method & External & \multicolumn{8}{|c}{BLEU} \\
 & Speech & \texttt{de} & \texttt{es} & \texttt{fr} & \texttt{it} & \texttt{nl} & \texttt{pt} & \texttt{ro} & \texttt{ru} \\
\hline
\hline
MTL \cite{9415058} & - & 23.9 & 28.6 & 33.1 & - & - & - & - & - \\
JT-S-MT \cite{tang-etal-2021-improving} & - & 26.8 & 31.0 & 37.4 & - & - & - & - & -\\
Chimera \cite{han-etal-2021-learning} & \checkmark & 27.1 & 30.6 & 35.6 & 25.0 & 29.2 & 30.2 & 24.0 & 17.4 \\
XSTNet \cite{ye21_interspeech} & \checkmark & 27.1 & 30.8 & 38.0 & 26.4 & 31.2 & 32.4 & 25.7 & 18.5 \\
STEMM \cite{fang-etal-2022-stemm} & \checkmark & 28.7 & 31.0 & 37.4 & 25.8 & 30.5 & 31.7 & 24.5 & 17.8 \\
ConST \cite{ye-etal-2022-cross} & \checkmark & 28.3 & 32.0 & 38.3 & 27.2 & 31.7 & 33.1 & 25.6 & 18.9 \\
SpeechUT \cite{zhang-etal-2022-speechut}$^{\dagger}$ & \checkmark & \bf 30.1 & \bf 33.6 & \bf 41.4 & - & - & - & - & - \\
WACO \cite{ouyang-etal-2023-waco} & \checkmark & 28.1 & 32.0 & 38.1 & - & - & - & - & - \\
M$^3$ST \cite{10095090} & \checkmark & 29.3 & 32.4 & 38.5 & 27.5 & 32.5 & 33.4 & 25.9 & 19.3 \\
FCCL$^{m}$ \cite{10042965} & \checkmark & 29.0 & 31.9 & 38.3 & 27.3 & 31.6 & 32.7 & 26.8 & 19.7 \\
CMOT \cite{zhou-etal-2023-cmot} & \checkmark & 29.0 & 32.8 & 39.5 & 27.5 & 32.1 & 33.5 & 26.0 & 19.2 \\
CRESS \cite{fang-feng-2023-understanding} & \checkmark & 29.4 & 33.2 & 40.1 & 27.6 & 32.3 & 33.6 & 26.4 & 19.7 \\
\hline
W2V2-Transformer & \checkmark & 27.3 & 31.7 & 38.0 & 26.3 & 29.8 & 31.7 & 23.4 & 18.2 \\
\ \ + SimRegCR$^-$ & \checkmark & 29.0 & 33.0 & 39.4 & 27.3 & 32.2 & 33.5 & 26.0 & 19.4 \\
\ \ + SimRegCR & \checkmark & $29.2^*$ & $33.0^*$ & $40.0^*$ & $\bf 28.2^*$ & $\bf 32.7^*$ & $\bf 34.2^*$ & $\bf 26.7^*$ & $\bf 20.1^*$ \\
\end{tabular}
\caption{Case-sensitive detokenized BLEU scores on MuST-C \texttt{tst-COMMON} set with external MT datasets. "External speech" denotes unlabeled speech data. $\dagger$ is a speech-unit-text pretraining model whose training costs are much higher than ours. * indicates the improvements over W2V2-Transformer are statistically significant with $p < 0.01$. The highest BLEU scores are marked in bold for all methods in each column.}\label{tab:res_with_mt}
\end{table*}


\subsection{Zero-shot End-to-End Speech Translation}\label{sec:zs_e2e_st}

The experimental results with external MT data are summarized in Table \ref{tab:res_with_mt}. For fair comparison, we keep our experimental settings consistent with \citet{wang-etal-2022-discrete} to use WMT14 dataset for \texttt{en}$\rightarrow$\texttt{de}/\texttt{es}/\texttt{fr}/\texttt{ru} as the external MT data\footnote{We only use europarl v7, commoncrawl, and news commentary subsets of WMT14 dataset for \texttt{en}$\rightarrow$\texttt{fr}.}. During inference, we use beam search decoding with a beam size of $5$ with length penalty $1.0$. The detailed information of the baseline methods are summarized in Appendix \ref{sec:zs_e2e_st_baseline}, and the corresponding BLEU scores are reported from \citet{wang-etal-2022-discrete}. The choice for hyperparameters and the corresponding model performance in each training step of our approach are summarized in Table \ref{tab:15}.

\begin{table}[h]\small
\centering
\begin{tabular}{l | c c c c}
Method & \multicolumn{4}{|c}{BLEU} \\
& \texttt{de} & \texttt{es} & \texttt{fr} & \texttt{ru} \\
\hline
\hline
MultiSLT & 6.8 & 6.8 & 10.9 & - \\
Chimera & 13.5 & 15.3 & 22.2 & 8.3 \\
DCMA & 24.0 & 26.2 & 33.1 & \bf 16.0 \\
\hline
W2V2-Transformer & 0.5 & 0.4 & 0.4 & 0.1 \\
\ \ + SimZeroCR & \bf 25.1 & \bf 27.0 & \bf 34.6 & 15.6 \\
\end{tabular}
\caption{Case-sensitive detokenized BLEU scores on MuST-C \texttt{tst-COMMON} set with external MT datasets. The highest BLEU scores are marked in bold for all methods in each column.}\label{tab:res_with_mt}
\end{table}

Despite the language tag is properly set during inference, W2V2-Transformer is still not capable of translating into specific language and only generating English text. We can see that SimZeroCR gains an average improvement of $25.2$ BLEU scores over the W2V2-Transformer baseline and achieves an average improvement of $0.8$ BLEU scores over the current SOTA method DCMA that incorporates shared memory and vector quantization modules, clearly showing the effectiveness of our method.

\section{Related Work}

E2E ST is a cross-modal task, and one major challenge is direct ST data scarcity. To address such problem, people usually adopt MT data by leveraging the techniques such as pretraining \cite{bansal-etal-2019-pre,alinejad-sarkar-2020-effectively,le-etal-2021-lightweight,tang-etal-2022-unified}, multi-task learning \cite{le-etal-2020-dual,dong2021listen,9414703}, knowledge distillation \cite{liu19d_interspeech,gaido-etal-2020-end,inaguma-etal-2021-source}, and data augmentation \cite{lam-etal-2022-sample,fang-feng-2023-back}. Due to the representation discrepancy between speech and text modalities, people also utilize cross-modal alignment \cite{han-etal-2021-learning,fang-etal-2022-stemm,ye-etal-2022-cross,ouyang-etal-2023-waco} to fully exploit MT data. Specifically, \citet{wang-etal-2022-discrete} employ a shared discrete vocabulary space to accommodate both modalities of speech and text and achieve SOTA performance in the zero-shot setting. We show that the zero-shot E2E ST performance could be boosted by leveraging simple cross-modal consistency regularization. \citet{fang-feng-2023-understanding} propose the cross-modal regularization with scheduled sampling method to bridge the modality gap and achieve the SOTA performance in the regular setting. We find that the regularization is more crucial than modality adaption, which is in line with \citet{han-etal-2023-modality}, and achieve the SOTA performance in the regular setting by leveraging simple intra-modal consistency regularization.

\section{Conclusion}\label{sec:conclusion}

In this paper, we propose two simple but effective consistency regularization based strategies for learning E2E ST models. We analyze the regularization effect of SimRegCR on the regular E2E ST performance and show that SimZeroCR could effectively close the modality gap. Experiments on the MuST-C benchmark demonstrate the capabilities of our approaches to improve translation performance in both regular and zero-shot settings. Given the universality and simplicity of SimRegCR and SimZeroCR, we believe they can serve as strong baselines for future E2E ST research. For future work, we will explore the effectiveness of consistency regularization on more speech related tasks, such as speech-to-speech translation, speech language modeling, etc.




\bibliography{anthology,custom}

\begin{thebibliography}{53}
\expandafter\ifx\csname natexlab\endcsname\relax\def\natexlab#1{#1}\fi

\bibitem[{Alinejad and Sarkar(2020)}]{alinejad-sarkar-2020-effectively}
Ashkan Alinejad and Anoop Sarkar. 2020.
\newblock \href {https://doi.org/10.18653/v1/2020.emnlp-main.644} {Effectively
  pretraining a speech translation decoder with machine translation data}.
\newblock In \emph{Proceedings of the 2020 Conference on Empirical Methods in
  Natural Language Processing (EMNLP)}, pages 8014--8020, Online. Association
  for Computational Linguistics.

\bibitem[{Baevski et~al.(2020)Baevski, Zhou, Mohamed, and
  Auli}]{NEURIPS2020_92d1e1eb}
Alexei Baevski, Yuhao Zhou, Abdelrahman Mohamed, and Michael Auli. 2020.
\newblock \href
  {https://proceedings.neurips.cc/paper/2020/hash/92d1e1eb1cd6f9fba3227870bb6d7f07-Abstract.html}
  {wav2vec 2.0: {A} framework for self-supervised learning of speech
  representations}.
\newblock In \emph{Advances in Neural Information Processing Systems 33: Annual
  Conference on Neural Information Processing Systems 2020, NeurIPS 2020,
  December 6-12, 2020, virtual}.

\bibitem[{Bansal et~al.(2019)Bansal, Kamper, Livescu, Lopez, and
  Goldwater}]{bansal-etal-2019-pre}
Sameer Bansal, Herman Kamper, Karen Livescu, Adam Lopez, and Sharon Goldwater.
  2019.
\newblock \href {https://doi.org/10.18653/v1/N19-1006} {Pre-training on
  high-resource speech recognition improves low-resource speech-to-text
  translation}.
\newblock In \emph{Proceedings of the 2019 Conference of the North {A}merican
  Chapter of the Association for Computational Linguistics: Human Language
  Technologies, Volume 1 (Long and Short Papers)}, pages 58--68, Minneapolis,
  Minnesota. Association for Computational Linguistics.

\bibitem[{Berard et~al.(2016)Berard, Pietquin, Servan, and
  Besacier}]{berard2016listen}
Alexandre Berard, Olivier Pietquin, Christophe Servan, and Laurent Besacier.
  2016.
\newblock \href {http://arxiv.org/abs/1612.01744} {Listen and translate: {A}
  proof of concept for end-to-end speech-to-text translation}.
\newblock \emph{CoRR}, abs/1612.01744.

\bibitem[{Bojar et~al.(2013)Bojar, Buck, Callison-Burch, Federmann, Haddow,
  Koehn, Monz, Post, Soricut, and Specia}]{bojar-etal-2013-findings}
Ond{\v{r}}ej Bojar, Christian Buck, Chris Callison-Burch, Christian Federmann,
  Barry Haddow, Philipp Koehn, Christof Monz, Matt Post, Radu Soricut, and
  Lucia Specia. 2013.
\newblock \href {https://aclanthology.org/W13-2201} {Findings of the 2013
  {W}orkshop on {S}tatistical {M}achine {T}ranslation}.
\newblock In \emph{Proceedings of the Eighth Workshop on Statistical Machine
  Translation}, pages 1--44, Sofia, Bulgaria. Association for Computational
  Linguistics.

\bibitem[{Bojar et~al.(2014)Bojar, Buck, Federmann, Haddow, Koehn, Leveling,
  Monz, Pecina, Post, Saint-Amand, Soricut, Specia, and
  Tamchyna}]{bojar-etal-2014-findings}
Ond{\v{r}}ej Bojar, Christian Buck, Christian Federmann, Barry Haddow, Philipp
  Koehn, Johannes Leveling, Christof Monz, Pavel Pecina, Matt Post, Herve
  Saint-Amand, Radu Soricut, Lucia Specia, and Ale{\v{s}} Tamchyna. 2014.
\newblock \href {https://doi.org/10.3115/v1/W14-3302} {Findings of the 2014
  workshop on statistical machine translation}.
\newblock In \emph{Proceedings of the Ninth Workshop on Statistical Machine
  Translation}, pages 12--58, Baltimore, Maryland, USA. Association for
  Computational Linguistics.

\bibitem[{Bojar et~al.(2016)Bojar, Chatterjee, Federmann, Graham, Haddow, Huck,
  Jimeno~Yepes, Koehn, Logacheva, Monz, Negri, N{\'e}v{\'e}ol, Neves, Popel,
  Post, Rubino, Scarton, Specia, Turchi, Verspoor, and
  Zampieri}]{bojar-etal-2016-findings}
Ond{\v{r}}ej Bojar, Rajen Chatterjee, Christian Federmann, Yvette Graham, Barry
  Haddow, Matthias Huck, Antonio Jimeno~Yepes, Philipp Koehn, Varvara
  Logacheva, Christof Monz, Matteo Negri, Aur{\'e}lie N{\'e}v{\'e}ol, Mariana
  Neves, Martin Popel, Matt Post, Raphael Rubino, Carolina Scarton, Lucia
  Specia, Marco Turchi, Karin Verspoor, and Marcos Zampieri. 2016.
\newblock \href {https://doi.org/10.18653/v1/W16-2301} {Findings of the 2016
  conference on machine translation}.
\newblock In \emph{Proceedings of the First Conference on Machine Translation:
  Volume 2, Shared Task Papers}, pages 131--198, Berlin, Germany. Association
  for Computational Linguistics.

\bibitem[{Chen et~al.(2021)Chen, Fan, Zhang, Chen, and
  Huang}]{chen-etal-2021-manifold}
Guandan Chen, Kai Fan, Kaibo Zhang, Boxing Chen, and Zhongqiang Huang. 2021.
\newblock \href {https://doi.org/10.18653/v1/2021.findings-acl.281} {Manifold
  adversarial augmentation for neural machine translation}.
\newblock In \emph{Findings of the Association for Computational Linguistics:
  ACL-IJCNLP 2021}, pages 3184--3189, Online. Association for Computational
  Linguistics.

\bibitem[{Cheng et~al.(2023)Cheng, Dong, Yue, Ko, Wang, and Zou}]{10095090}
Xuxin Cheng, Qianqian Dong, Fengpeng Yue, Tom Ko, Mingxuan Wang, and Yuexian
  Zou. 2023.
\newblock \href {https://doi.org/10.1109/ICASSP49357.2023.10095090} {M3st: Mix
  at three levels for speech translation}.
\newblock In \emph{ICASSP 2023 - 2023 IEEE International Conference on
  Acoustics, Speech and Signal Processing (ICASSP)}, pages 1--5.

\bibitem[{Di~Gangi et~al.(2019)Di~Gangi, Cattoni, Bentivogli, Negri, and
  Turchi}]{di-gangi-etal-2019-must}
Mattia~A. Di~Gangi, Roldano Cattoni, Luisa Bentivogli, Matteo Negri, and Marco
  Turchi. 2019.
\newblock \href {https://doi.org/10.18653/v1/N19-1202} {{M}u{ST}-{C}: a
  {M}ultilingual {S}peech {T}ranslation {C}orpus}.
\newblock In \emph{Proceedings of the 2019 Conference of the North {A}merican
  Chapter of the Association for Computational Linguistics: Human Language
  Technologies, Volume 1 (Long and Short Papers)}, pages 2012--2017,
  Minneapolis, Minnesota. Association for Computational Linguistics.

\bibitem[{Dong et~al.(2021)Dong, Ye, Wang, Zhou, Xu, Xu, and
  Li}]{dong2021listen}
Qianqian Dong, Rong Ye, Mingxuan Wang, Hao Zhou, Shuang Xu, Bo~Xu, and Lei Li.
  2021.
\newblock Listen, understand and translate: Triple supervision decouples
  end-to-end speech-to-text translation.
\newblock In \emph{Proceedings of the AAAI Conference on Artificial
  Intelligence}, volume~35, pages 12749--12759.

\bibitem[{Duong et~al.(2016)Duong, Anastasopoulos, Chiang, Bird, and
  Cohn}]{duong-etal-2016-attentional}
Long Duong, Antonios Anastasopoulos, David Chiang, Steven Bird, and Trevor
  Cohn. 2016.
\newblock \href {https://doi.org/10.18653/v1/N16-1109} {An attentional model
  for speech translation without transcription}.
\newblock In \emph{Proceedings of the 2016 Conference of the North {A}merican
  Chapter of the Association for Computational Linguistics: Human Language
  Technologies}, pages 949--959, San Diego, California. Association for
  Computational Linguistics.

\bibitem[{Escolano et~al.(2021)Escolano, Costa-jussà, Fonollosa, and
  Segura}]{9688026}
Carlos Escolano, Marta~R. Costa-jussà, José A.~R. Fonollosa, and Carlos
  Segura. 2021.
\newblock \href {https://doi.org/10.1109/ASRU51503.2021.9688026} {Enabling
  zero-shot multilingual spoken language translation with language-specific
  encoders and decoders}.
\newblock In \emph{2021 IEEE Automatic Speech Recognition and Understanding
  Workshop (ASRU)}, pages 694--701.

\bibitem[{Fang and Feng(2023{\natexlab{a}})}]{fang-feng-2023-back}
Qingkai Fang and Yang Feng. 2023{\natexlab{a}}.
\newblock \href {https://aclanthology.org/2023.acl-long.251} {Back translation
  for speech-to-text translation without transcripts}.
\newblock In \emph{Proceedings of the 61st Annual Meeting of the Association
  for Computational Linguistics (Volume 1: Long Papers)}, pages 4567--4587,
  Toronto, Canada. Association for Computational Linguistics.

\bibitem[{Fang and Feng(2023{\natexlab{b}})}]{fang-feng-2023-understanding}
Qingkai Fang and Yang Feng. 2023{\natexlab{b}}.
\newblock \href {https://aclanthology.org/2023.acl-long.884} {Understanding and
  bridging the modality gap for speech translation}.
\newblock In \emph{Proceedings of the 61st Annual Meeting of the Association
  for Computational Linguistics (Volume 1: Long Papers)}, pages 15864--15881,
  Toronto, Canada. Association for Computational Linguistics.

\bibitem[{Fang et~al.(2022)Fang, Ye, Li, Feng, and Wang}]{fang-etal-2022-stemm}
Qingkai Fang, Rong Ye, Lei Li, Yang Feng, and Mingxuan Wang. 2022.
\newblock \href {https://doi.org/10.18653/v1/2022.acl-long.486} {{STEMM}:
  Self-learning with speech-text manifold mixup for speech translation}.
\newblock In \emph{Proceedings of the 60th Annual Meeting of the Association
  for Computational Linguistics (Volume 1: Long Papers)}, pages 7050--7062,
  Dublin, Ireland. Association for Computational Linguistics.

\bibitem[{Gaido et~al.(2020)Gaido, Di~Gangi, Negri, and
  Turchi}]{gaido-etal-2020-end}
Marco Gaido, Mattia~A. Di~Gangi, Matteo Negri, and Marco Turchi. 2020.
\newblock \href {https://doi.org/10.18653/v1/2020.iwslt-1.8} {End-to-end
  speech-translation with knowledge distillation: {FBK}@{IWSLT}2020}.
\newblock In \emph{Proceedings of the 17th International Conference on Spoken
  Language Translation}, pages 80--88, Online. Association for Computational
  Linguistics.

\bibitem[{Gao et~al.(2022)Gao, He, Wu, and Wang}]{gao-etal-2022-bi}
Pengzhi Gao, Zhongjun He, Hua Wu, and Haifeng Wang. 2022.
\newblock \href {https://doi.org/10.18653/v1/2022.naacl-main.289}
  {{B}i-{S}im{C}ut: A simple strategy for boosting neural machine translation}.
\newblock In \emph{Proceedings of the 2022 Conference of the North American
  Chapter of the Association for Computational Linguistics: Human Language
  Technologies}, pages 3938--3948, Seattle, United States. Association for
  Computational Linguistics.

\bibitem[{Gao et~al.(2023)Gao, Zhang, He, Wu, and
  Wang}]{gao-etal-2023-improving}
Pengzhi Gao, Liwen Zhang, Zhongjun He, Hua Wu, and Haifeng Wang. 2023.
\newblock \href {https://aclanthology.org/2023.findings-acl.766} {Improving
  zero-shot multilingual neural machine translation by leveraging cross-lingual
  consistency regularization}.
\newblock In \emph{Findings of the Association for Computational Linguistics:
  ACL 2023}, pages 12103--12119, Toronto, Canada. Association for Computational
  Linguistics.

\bibitem[{Han et~al.(2021)Han, Wang, Ji, and Li}]{han-etal-2021-learning}
Chi Han, Mingxuan Wang, Heng Ji, and Lei Li. 2021.
\newblock \href {https://doi.org/10.18653/v1/2021.findings-acl.195} {Learning
  shared semantic space for speech-to-text translation}.
\newblock In \emph{Findings of the Association for Computational Linguistics:
  ACL-IJCNLP 2021}, pages 2214--2225, Online. Association for Computational
  Linguistics.

\bibitem[{Han et~al.(2023)Han, Xu, Xiao, and Zhu}]{han-etal-2023-modality}
Yuchen Han, Chen Xu, Tong Xiao, and Jingbo Zhu. 2023.
\newblock \href {https://aclanthology.org/2023.acl-short.115} {Modality
  adaption or regularization? a case study on end-to-end speech translation}.
\newblock In \emph{Proceedings of the 61st Annual Meeting of the Association
  for Computational Linguistics (Volume 2: Short Papers)}, pages 1340--1348,
  Toronto, Canada. Association for Computational Linguistics.

\bibitem[{Hinton and Roweis(2002)}]{NIPS2002_6150ccc6}
Geoffrey~E Hinton and Sam Roweis. 2002.
\newblock \href
  {https://proceedings.neurips.cc/paper_files/paper/2002/file/6150ccc6069bea6b5716254057a194ef-Paper.pdf}
  {Stochastic neighbor embedding}.
\newblock In \emph{Advances in Neural Information Processing Systems},
  volume~15. MIT Press.

\bibitem[{Hsu et~al.(2021)Hsu, Bolte, Tsai, Lakhotia, Salakhutdinov, and
  Mohamed}]{10.1109/TASLP.2021.3122291}
Wei-Ning Hsu, Benjamin Bolte, Yao-Hung~Hubert Tsai, Kushal Lakhotia, Ruslan
  Salakhutdinov, and Abdelrahman Mohamed. 2021.
\newblock \href {https://doi.org/10.1109/TASLP.2021.3122291} {Hubert:
  Self-supervised speech representation learning by masked prediction of hidden
  units}.
\newblock \emph{IEEE/ACM Trans. Audio, Speech and Lang. Proc.}, 29:3451–3460.

\bibitem[{Inaguma et~al.(2021)Inaguma, Kawahara, and
  Watanabe}]{inaguma-etal-2021-source}
Hirofumi Inaguma, Tatsuya Kawahara, and Shinji Watanabe. 2021.
\newblock \href {https://doi.org/10.18653/v1/2021.naacl-main.150} {Source and
  target bidirectional knowledge distillation for end-to-end speech
  translation}.
\newblock In \emph{Proceedings of the 2021 Conference of the North American
  Chapter of the Association for Computational Linguistics: Human Language
  Technologies}, pages 1872--1881, Online. Association for Computational
  Linguistics.

\bibitem[{Indurthi et~al.(2021)Indurthi, Zaidi, Kumar~Lakumarapu, Lee, Han,
  Ahn, Kim, Kim, and Hwang}]{9414703}
Sathish Indurthi, Mohd~Abbas Zaidi, Nikhil Kumar~Lakumarapu, Beomseok Lee,
  Hyojung Han, Seokchan Ahn, Sangha Kim, Chanwoo Kim, and Inchul Hwang. 2021.
\newblock \href {https://doi.org/10.1109/ICASSP39728.2021.9414703} {Task aware
  multi-task learning for speech to text tasks}.
\newblock In \emph{ICASSP 2021 - 2021 IEEE International Conference on
  Acoustics, Speech and Signal Processing (ICASSP)}, pages 7723--7727.

\bibitem[{Kudo and Richardson(2018)}]{kudo-richardson-2018-sentencepiece}
Taku Kudo and John Richardson. 2018.
\newblock \href {https://doi.org/10.18653/v1/D18-2012} {{S}entence{P}iece: A
  simple and language independent subword tokenizer and detokenizer for neural
  text processing}.
\newblock In \emph{Proceedings of the 2018 Conference on Empirical Methods in
  Natural Language Processing: System Demonstrations}, pages 66--71, Brussels,
  Belgium. Association for Computational Linguistics.

\bibitem[{Lam et~al.(2022)Lam, Schamoni, and Riezler}]{lam-etal-2022-sample}
Tsz~Kin Lam, Shigehiko Schamoni, and Stefan Riezler. 2022.
\newblock \href {https://doi.org/10.18653/v1/2022.acl-short.27} {Sample,
  translate, recombine: Leveraging audio alignments for data augmentation in
  end-to-end speech translation}.
\newblock In \emph{Proceedings of the 60th Annual Meeting of the Association
  for Computational Linguistics (Volume 2: Short Papers)}, pages 245--254,
  Dublin, Ireland. Association for Computational Linguistics.

\bibitem[{Le et~al.(2020)Le, Pino, Wang, Gu, Schwab, and
  Besacier}]{le-etal-2020-dual}
Hang Le, Juan Pino, Changhan Wang, Jiatao Gu, Didier Schwab, and Laurent
  Besacier. 2020.
\newblock \href {https://doi.org/10.18653/v1/2020.coling-main.314}
  {Dual-decoder transformer for joint automatic speech recognition and
  multilingual speech translation}.
\newblock In \emph{Proceedings of the 28th International Conference on
  Computational Linguistics}, pages 3520--3533, Barcelona, Spain (Online).
  International Committee on Computational Linguistics.

\bibitem[{Le et~al.(2021)Le, Pino, Wang, Gu, Schwab, and
  Besacier}]{le-etal-2021-lightweight}
Hang Le, Juan Pino, Changhan Wang, Jiatao Gu, Didier Schwab, and Laurent
  Besacier. 2021.
\newblock \href {https://doi.org/10.18653/v1/2021.acl-short.103} {Lightweight
  adapter tuning for multilingual speech translation}.
\newblock In \emph{Proceedings of the 59th Annual Meeting of the Association
  for Computational Linguistics and the 11th International Joint Conference on
  Natural Language Processing (Volume 2: Short Papers)}, pages 817--824,
  Online. Association for Computational Linguistics.

\bibitem[{Liang et~al.(2021)Liang, Wu, Li, Wang, Meng, Qin, Chen, Zhang, and
  Liu}]{NEURIPS2021_5a66b920}
Xiaobo Liang, Lijun Wu, Juntao Li, Yue Wang, Qi~Meng, Tao Qin, Wei Chen, Min
  Zhang, and Tie-Yan Liu. 2021.
\newblock \href
  {https://proceedings.neurips.cc/paper_files/paper/2021/file/5a66b9200f29ac3fa0ae244cc2a51b39-Paper.pdf}
  {R-drop: Regularized dropout for neural networks}.
\newblock In \emph{Advances in Neural Information Processing Systems},
  volume~34, pages 10890--10905. Curran Associates, Inc.

\bibitem[{Liu et~al.(2019)Liu, Xiong, Zhang, He, Wu, Wang, and
  Zong}]{liu19d_interspeech}
Yuchen Liu, Hao Xiong, Jiajun Zhang, Zhongjun He, Hua Wu, Haifeng Wang, and
  Chengqing Zong. 2019.
\newblock \href {https://doi.org/10.21437/Interspeech.2019-2582} {{End-to-End
  Speech Translation with Knowledge Distillation}}.
\newblock In \emph{Proc. Interspeech 2019}, pages 1128--1132.

\bibitem[{Ouyang et~al.(2023)Ouyang, Ye, and Li}]{ouyang-etal-2023-waco}
Siqi Ouyang, Rong Ye, and Lei Li. 2023.
\newblock \href {https://aclanthology.org/2023.acl-long.216} {{WACO}:
  Word-aligned contrastive learning for speech translation}.
\newblock In \emph{Proceedings of the 61st Annual Meeting of the Association
  for Computational Linguistics (Volume 1: Long Papers)}, pages 3891--3907,
  Toronto, Canada. Association for Computational Linguistics.

\bibitem[{Panayotov et~al.(2015)Panayotov, Chen, Povey, and
  Khudanpur}]{panayotov2015librispeech}
Vassil Panayotov, Guoguo Chen, Daniel Povey, and Sanjeev Khudanpur. 2015.
\newblock \href {https://doi.org/10.1109/ICASSP.2015.7178964} {Librispeech: An
  {ASR} corpus based on public domain audio books}.
\newblock In \emph{2015 {IEEE} International Conference on Acoustics, Speech
  and Signal Processing, {ICASSP} 2015, South Brisbane, Queensland, Australia,
  April 19-24, 2015}, pages 5206--5210. {IEEE}.

\bibitem[{Papi et~al.(2021)Papi, Gaido, Negri, and
  Turchi}]{papi-etal-2021-speechformer}
Sara Papi, Marco Gaido, Matteo Negri, and Marco Turchi. 2021.
\newblock \href {https://doi.org/10.18653/v1/2021.emnlp-main.127}
  {Speechformer: Reducing information loss in direct speech translation}.
\newblock In \emph{Proceedings of the 2021 Conference on Empirical Methods in
  Natural Language Processing}, pages 1698--1706, Online and Punta Cana,
  Dominican Republic. Association for Computational Linguistics.

\bibitem[{Post(2018)}]{post-2018-call}
Matt Post. 2018.
\newblock \href {https://doi.org/10.18653/v1/W18-6319} {A call for clarity in
  reporting {BLEU} scores}.
\newblock In \emph{Proceedings of the Third Conference on Machine Translation:
  Research Papers}, pages 186--191, Brussels, Belgium. Association for
  Computational Linguistics.

\bibitem[{Sato et~al.(2019)Sato, Suzuki, and Kiyono}]{sato-etal-2019-effective}
Motoki Sato, Jun Suzuki, and Shun Kiyono. 2019.
\newblock \href {https://doi.org/10.18653/v1/P19-1020} {Effective adversarial
  regularization for neural machine translation}.
\newblock In \emph{Proceedings of the 57th Annual Meeting of the Association
  for Computational Linguistics}, pages 204--210, Florence, Italy. Association
  for Computational Linguistics.

\bibitem[{Sperber et~al.(2017)Sperber, Neubig, Niehues, and
  Waibel}]{sperber-etal-2017-neural}
Matthias Sperber, Graham Neubig, Jan Niehues, and Alex Waibel. 2017.
\newblock \href {https://doi.org/10.18653/v1/D17-1145} {Neural
  lattice-to-sequence models for uncertain inputs}.
\newblock In \emph{Proceedings of the 2017 Conference on Empirical Methods in
  Natural Language Processing}, pages 1380--1389, Copenhagen, Denmark.
  Association for Computational Linguistics.

\bibitem[{Sperber et~al.(2019)Sperber, Neubig, Pham, and
  Waibel}]{sperber-etal-2019-self}
Matthias Sperber, Graham Neubig, Ngoc-Quan Pham, and Alex Waibel. 2019.
\newblock \href {https://doi.org/10.18653/v1/P19-1115} {Self-attentional models
  for lattice inputs}.
\newblock In \emph{Proceedings of the 57th Annual Meeting of the Association
  for Computational Linguistics}, pages 1185--1197, Florence, Italy.
  Association for Computational Linguistics.

\bibitem[{Tang et~al.(2022)Tang, Gong, Dong, Wang, Hsu, Gu, Baevski, Li,
  Mohamed, Auli, and Pino}]{tang-etal-2022-unified}
Yun Tang, Hongyu Gong, Ning Dong, Changhan Wang, Wei-Ning Hsu, Jiatao Gu,
  Alexei Baevski, Xian Li, Abdelrahman Mohamed, Michael Auli, and Juan Pino.
  2022.
\newblock \href {https://doi.org/10.18653/v1/2022.acl-long.105} {Unified
  speech-text pre-training for speech translation and recognition}.
\newblock In \emph{Proceedings of the 60th Annual Meeting of the Association
  for Computational Linguistics (Volume 1: Long Papers)}, pages 1488--1499,
  Dublin, Ireland. Association for Computational Linguistics.

\bibitem[{Tang et~al.(2021{\natexlab{a}})Tang, Pino, Li, Wang, and
  Genzel}]{tang-etal-2021-improving}
Yun Tang, Juan Pino, Xian Li, Changhan Wang, and Dmitriy Genzel.
  2021{\natexlab{a}}.
\newblock \href {https://doi.org/10.18653/v1/2021.acl-long.328} {Improving
  speech translation by understanding and learning from the auxiliary text
  translation task}.
\newblock In \emph{Proceedings of the 59th Annual Meeting of the Association
  for Computational Linguistics and the 11th International Joint Conference on
  Natural Language Processing (Volume 1: Long Papers)}, pages 4252--4261,
  Online. Association for Computational Linguistics.

\bibitem[{Tang et~al.(2021{\natexlab{b}})Tang, Pino, Wang, Ma, and
  Genzel}]{9415058}
Yun Tang, Juan Pino, Changhan Wang, Xutai Ma, and Dmitriy Genzel.
  2021{\natexlab{b}}.
\newblock \href {https://doi.org/10.1109/ICASSP39728.2021.9415058} {A general
  multi-task learning framework to leverage text data for speech to text
  tasks}.
\newblock In \emph{ICASSP 2021 - 2021 IEEE International Conference on
  Acoustics, Speech and Signal Processing (ICASSP)}, pages 6209--6213.

\bibitem[{Vaswani et~al.(2017)Vaswani, Shazeer, Parmar, Uszkoreit, Jones,
  Gomez, Kaiser, and Polosukhin}]{vaswani2017attention}
Ashish Vaswani, Noam Shazeer, Niki Parmar, Jakob Uszkoreit, Llion Jones,
  Aidan~N. Gomez, Lukasz Kaiser, and Illia Polosukhin. 2017.
\newblock \href
  {https://proceedings.neurips.cc/paper/2017/hash/3f5ee243547dee91fbd053c1c4a845aa-Abstract.html}
  {Attention is all you need}.
\newblock In \emph{Advances in Neural Information Processing Systems 30: Annual
  Conference on Neural Information Processing Systems 2017, December 4-9, 2017,
  Long Beach, CA, {USA}}, pages 5998--6008.

\bibitem[{Wang et~al.(2020{\natexlab{a}})Wang, Tang, Ma, Wu, Okhonko, and
  Pino}]{wang-etal-2020-fairseq}
Changhan Wang, Yun Tang, Xutai Ma, Anne Wu, Dmytro Okhonko, and Juan Pino.
  2020{\natexlab{a}}.
\newblock \href {https://aclanthology.org/2020.aacl-demo.6} {Fairseq {S}2{T}:
  Fast speech-to-text modeling with fairseq}.
\newblock In \emph{Proceedings of the 1st Conference of the Asia-Pacific
  Chapter of the Association for Computational Linguistics and the 10th
  International Joint Conference on Natural Language Processing: System
  Demonstrations}, pages 33--39, Suzhou, China. Association for Computational
  Linguistics.

\bibitem[{Wang et~al.(2022)Wang, Liu, Chen, Zhang, Luo, Huang, and
  Zong}]{wang-etal-2022-discrete}
Chen Wang, Yuchen Liu, Boxing Chen, Jiajun Zhang, Wei Luo, Zhongqiang Huang,
  and Chengqing Zong. 2022.
\newblock \href {https://aclanthology.org/2022.emnlp-main.354} {Discrete
  cross-modal alignment enables zero-shot speech translation}.
\newblock In \emph{Proceedings of the 2022 Conference on Empirical Methods in
  Natural Language Processing}, pages 5291--5302, Abu Dhabi, United Arab
  Emirates. Association for Computational Linguistics.

\bibitem[{Wang et~al.(2020{\natexlab{b}})Wang, Wu, Liu, Zhou, and
  Yang}]{wang-etal-2020-curriculum}
Chengyi Wang, Yu~Wu, Shujie Liu, Ming Zhou, and Zhenglu Yang.
  2020{\natexlab{b}}.
\newblock \href {https://doi.org/10.18653/v1/2020.acl-main.344} {Curriculum
  pre-training for end-to-end speech translation}.
\newblock In \emph{Proceedings of the 58th Annual Meeting of the Association
  for Computational Linguistics}, pages 3728--3738, Online. Association for
  Computational Linguistics.

\bibitem[{Xu et~al.(2021)Xu, Hu, Li, Zhang, Huang, Ju, Xiao, and
  Zhu}]{xu-etal-2021-stacked}
Chen Xu, Bojie Hu, Yanyang Li, Yuhao Zhang, Shen Huang, Qi~Ju, Tong Xiao, and
  Jingbo Zhu. 2021.
\newblock \href {https://doi.org/10.18653/v1/2021.acl-long.204} {Stacked
  acoustic-and-textual encoding: Integrating the pre-trained models into speech
  translation encoders}.
\newblock In \emph{Proceedings of the 59th Annual Meeting of the Association
  for Computational Linguistics and the 11th International Joint Conference on
  Natural Language Processing (Volume 1: Long Papers)}, pages 2619--2630,
  Online. Association for Computational Linguistics.

\bibitem[{Ye et~al.(2021)Ye, Wang, and Li}]{ye21_interspeech}
Rong Ye, Mingxuan Wang, and Lei Li. 2021.
\newblock \href {https://doi.org/10.21437/Interspeech.2021-1065} {{End-to-End
  Speech Translation via Cross-Modal Progressive Training}}.
\newblock In \emph{Proc. Interspeech 2021}, pages 2267--2271.

\bibitem[{Ye et~al.(2022)Ye, Wang, and Li}]{ye-etal-2022-cross}
Rong Ye, Mingxuan Wang, and Lei Li. 2022.
\newblock \href {https://doi.org/10.18653/v1/2022.naacl-main.376} {Cross-modal
  contrastive learning for speech translation}.
\newblock In \emph{Proceedings of the 2022 Conference of the North American
  Chapter of the Association for Computational Linguistics: Human Language
  Technologies}, pages 5099--5113, Seattle, United States. Association for
  Computational Linguistics.

\bibitem[{Zhang et~al.(2020)Zhang, Williams, Titov, and
  Sennrich}]{zhang-etal-2020-improving}
Biao Zhang, Philip Williams, Ivan Titov, and Rico Sennrich. 2020.
\newblock \href {https://doi.org/10.18653/v1/2020.acl-main.148} {Improving
  massively multilingual neural machine translation and zero-shot translation}.
\newblock In \emph{Proceedings of the 58th Annual Meeting of the Association
  for Computational Linguistics}, pages 1628--1639, Online. Association for
  Computational Linguistics.

\bibitem[{Zhang et~al.(2023)Zhang, Si, Chen, Zhang, Yang, Qu, and
  Zhang}]{10042965}
Hao Zhang, Nianwen Si, Yaqi Chen, Wenlin Zhang, Xukui Yang, Dan Qu, and
  Wei-Qiang Zhang. 2023.
\newblock \href {https://doi.org/10.1109/TASLP.2023.3244521} {Improving speech
  translation by cross-modal multi-grained contrastive learning}.
\newblock \emph{IEEE/ACM Transactions on Audio, Speech, and Language
  Processing}, 31:1075--1086.

\bibitem[{Zhang et~al.(2019)Zhang, Ge, Chen, and Fan}]{zhang-etal-2019-lattice}
Pei Zhang, Niyu Ge, Boxing Chen, and Kai Fan. 2019.
\newblock \href {https://doi.org/10.18653/v1/P19-1649} {Lattice transformer for
  speech translation}.
\newblock In \emph{Proceedings of the 57th Annual Meeting of the Association
  for Computational Linguistics}, pages 6475--6484, Florence, Italy.
  Association for Computational Linguistics.

\bibitem[{Zhang et~al.(2022)Zhang, Zhou, Ao, Liu, Dai, Li, and
  Wei}]{zhang-etal-2022-speechut}
Ziqiang Zhang, Long Zhou, Junyi Ao, Shujie Liu, Lirong Dai, Jinyu Li, and Furu
  Wei. 2022.
\newblock \href {https://aclanthology.org/2022.emnlp-main.108} {{S}peech{UT}:
  Bridging speech and text with hidden-unit for encoder-decoder based
  speech-text pre-training}.
\newblock In \emph{Proceedings of the 2022 Conference on Empirical Methods in
  Natural Language Processing}, pages 1663--1676, Abu Dhabi, United Arab
  Emirates. Association for Computational Linguistics.

\bibitem[{Zhou et~al.(2023)Zhou, Fang, and Feng}]{zhou-etal-2023-cmot}
Yan Zhou, Qingkai Fang, and Yang Feng. 2023.
\newblock \href {https://aclanthology.org/2023.acl-long.436} {{CMOT}:
  Cross-modal mixup via optimal transport for speech translation}.
\newblock In \emph{Proceedings of the 61st Annual Meeting of the Association
  for Computational Linguistics (Volume 1: Long Papers)}, pages 7873--7887,
  Toronto, Canada. Association for Computational Linguistics.

\end{thebibliography}
\bibliographystyle{acl_natbib}

\appendix

\section*{Appendix}

\section{Statistics of All Datasets}

\begin{table}[h]
\centering
\begin{tabular}{c | c c | c c} 
 & \multicolumn{2}{c|}{MuST-C} & \multicolumn{2}{c}{External MT} \\
\texttt{en}$\rightarrow$ & hours & \#sents & name & \#sents \\
\hline
\hline
\texttt{de} & 408 & 234K & WMT16 & 4.6M \\
\texttt{es} & 504 & 270K & WMT13 & 15.2M \\
\texttt{fr} & 492 & 292K & WMT14 & 40.8M \\
\texttt{it} & 465 & 258K & OPUS100 & 1.0M \\
\texttt{nl} & 442 & 253K & OPUS100 & 1.0M \\
\texttt{pt} & 385 & 211K & OPUS100 & 1.0M \\
\texttt{ro} & 432 & 240K & WMT16 & 0.6M \\
\texttt{ru} & 489 & 270K & WMT16 & 2.5M \\
\hline
\end{tabular}
\caption{Statistics of all datasets. \#sents refers to the number of parallel sentence pairs.}
\label{dataset_statics}
\end{table}

\section{The Choice for Hyperparameters in Tables \ref{tab:es_regular_st} and \ref{tab:es_zeroshot_st}}

\begin{table}[h]
\centering
\begin{tabular}{c | c c | c | c c} 
ID & $\alpha$ & $\beta$ & ID & $\alpha$ & $\beta$ \\
\hline
\hline
\circled{1} & - & - & \circled{2} & 5.0 & - \\
\circled{3} & - & - & \circled{4} & 5.0 & - \\
\circled{5} & - & - & \circled{6} & 5.0 & - \\
\circled{7} & 4.0 & - & \circled{8} & - & - \\
\circled{9} & - & - & \circled{10} & - & 5.0 \\
\circled{11} & 3.0 & - & \circled{12} & 3.0 & 5.0 \\
\circled{13} & - & - & \circled{14} & 0.5 & - \\
\circled{15} & - & - & \circled{16} & 1.0 & - \\
\circled{17} & - & - & \circled{18} & 3.0 & - \\
\circled{19} & 3.0 & - & \\
\hline
\end{tabular}
\caption{The choice for hyperparameters in Table \ref{tab:es_regular_st}.}
\label{tab:es_regular_st_hyperparameter}
\end{table}

\begin{table}[h]
\centering
\begin{tabular}{c | c c | c | c c} 
ID & $\alpha$ & $\beta$ & ID & $\alpha$ & $\beta$ \\
\hline
\hline
\circled{1} & - & - & \circled{2} & 0.5 & - \\
\circled{3} & - & - & \circled{4} & 1.0 & - \\
\circled{5} & - & - & \circled{6} & - & 45.0 \\
\circled{7} & 2.0 & - & \circled{8} & 2.0 & 120.0 \\
\hline
\end{tabular}
\caption{The choice for hyperparameters in Table \ref{tab:es_zeroshot_st}.}
\label{tab:es_zeroshot_st_hyperparameter}
\end{table}

\section{Regular E2E ST Methods}\label{sec:reg_e2e_st_baseline}

We compare our approach with the following methods on the MuST-C benchmark:

\begin{itemize}[leftmargin=*]

\item {\bf Fairseq ST} \cite{wang-etal-2020-fairseq}: Fairseq ST is a fairseq extension\footnote{\url{https://github.com/facebookresearch/fairseq/tree/main/examples/speech_to_text}} for speech-to-text modeling tasks such as speech translation, which includes end-to-end workflows and SOTA models with scalability and extensibility design.

\item {\bf Dual Decoder} \cite{le-etal-2020-dual}: This paper introduces a dual-decoder Transformer architecture for synchronous speech recognition and multilingual speech translation. 

\item {\bf Speechformer} \cite{papi-etal-2021-speechformer}: This paper introduces a Transformer-based ST model that able to encode the whole raw audio features without any sub-optimal initial sub-sampling.

\item {\bf SATE} \cite{xu-etal-2021-stacked}: This paper proposes a stacked acoustic-and-textual encoding method, which is straightforward to incorporate the pretrained models into ST. 

\item {\bf BiKD} \cite{inaguma-etal-2021-source}: To fully leverage knowledge in both source and target language directions for bilingual E2E ST models, this paper proposes bidirectional sequence-level knowledge distillation, in which both forward sequence-level knowledge distillation from a source-to-target NMT model and backward sequence-level knowledge distillation from a target-to-source NMT model are combined.

\item {\bf XSTNet} \cite{ye21_interspeech}: This paper proposes cross speech-text network, an extremely concise model which can accept bi-modal inputs and jointly train ST, ASR and MT tasks.

\item {\bf MTL} \cite{9415058}: This paper proposes a general multi-task learning framework to leverage text data for ASR and ST tasks.

\item {\bf JT-S-MT} \cite{tang-etal-2021-improving}: This paper proposes three techniques to increase knowledge transfer from the MT task to the ST task, which include parameter sharing and initialization strategy to improve the information sharing between tasks, cross-attentive regularization and online knowledge distillation to encourage the ST system to learn more from the auxiliary MT task and then generate similar model representations from different modalities.

\item {\bf STEMM} \cite{fang-etal-2022-stemm}: This paper proposes a speech-text manifold mixup method to mix up the speech representation sequences and word embedding sequences.

\item {\bf ConST} \cite{ye-etal-2022-cross}: This paper proposes a simple yet effective contrastive learning framework bridging the speech-text representation gap and facilitating the ST with limited data.

\item {\bf SpeechUT} \cite{zhang-etal-2022-speechut}: This paper proposes a unified-modal speech-unit-text pretraining model, which bridges the modality gap between speech and text representations with hidden units.

\item {\bf WACO} \cite{ouyang-etal-2023-waco}: This paper proposes a simple and effective method for extremely low-resource speech-to-text translation, where the key idea is bridging word-level representations for both speech and text modalities via contrastive learning.

\item {\bf M$^3$ST} \cite{10095090}: This paper proposes a method to mix the training corpus at three levels, including word level, sentence level and frame level.

\item {\bf FCCL$^{m}$} \cite{10042965}: This paper proposes a cross-modal multi-grained contrast learning method for explicit knowledge transfer from the MT to the ST model. 

\item {\bf CMOT} \cite{zhou-etal-2023-cmot}: This paper proposes cross-modal mixup via optimal transport to adaptively find the alignment between speech and text sequences, and to mix up the sequences of different modalities at the token level.

\item {\bf CRESS} \cite{fang-feng-2023-understanding}: This paper proposes a simple yet effective method to regularize the model predictions of ST and MT, whose target-side contexts contain both ground truth words and self-generated words with scheduled sampling.

\end{itemize}

\section{Zero-shot E2E ST Methods}\label{sec:zs_e2e_st_baseline}

We compare our approach with the following methods on the MuST-C benchmark:

\begin{itemize}[leftmargin=*]

\item {\bf MultiSLT} \cite{9688026}: This paper extends the multilingual NMT system to perform spoken language translation and zero-shot multilingual spoken language translation by coupling language-specific encoder-decoders, even from monolingual ASR data only.

\item {\bf Chimera} \cite{han-etal-2021-learning}: This paper proposes a model capable of learning a text-speech shared semantic memory network for bridging the gap between speech and text representations.

\item {\bf DCMA} \cite{wang-etal-2022-discrete}: This paper proposes an alignment method to enable zero-shot ST, where the key part is to discretize the continuous vectors to a finite set of virtual tokens and use ASR data to map the corresponding speech and text to the same virtual token in the shared codebook.

\end{itemize}

\section{The Choice for Hyperparameters in Section \ref{sec:exps}}

\begin{table*}[h]
\centering
\begin{tabular}{c | c | c c c c c c c c} 
Training Stage & & \texttt{de} & \texttt{es} & \texttt{fr} & \texttt{it} & \texttt{nl} & \texttt{pt} & \texttt{ro} & \texttt{ru} \\
\hline
\hline
MT pretrain & Baseline & 29.33 & 34.61 & 41.47 & 31.25 & 34.41 & 35.80 & 28.13 & 19.40 \\
\hline
\multirow{3}{*}{ST finetune} & Baseline & 24.38 & 29.92 & 34.73 & 25.13 & 29.29 & 30.32 & 23.39 & 16.45 \\
\cline{2-10}
& BLEU & 27.35 & 31.53 & 38.10 & 27.24 & 32.00 & 33.30 & 25.89 & 18.83 \\
& $\alpha$ & 5 & 4 & 4 & 5 & 4 & 5 & 4 & 4 \\
\hline
\end{tabular}
\caption{The choice for hyperparameters and the corresponding MT \& ST performance in the training steps of SimRegCR$^-$ without external MT datasets.}
\label{tab:11}
\end{table*}

\begin{table*}[h]
\centering
\begin{tabular}{c | c | c c c c c c c c} 
Training Stage & & \texttt{de} & \texttt{es} & \texttt{fr} & \texttt{it} & \texttt{nl} & \texttt{pt} & \texttt{ro} & \texttt{ru} \\
\hline
\hline
\multirow{2}{*}{MT pretrain} & BLEU & 32.76 & 37.10 & 45.68 & 33.31 & 37.89 & 39.12 & 31.60 & 21.60 \\
& $\alpha$ & 5 & 5 & 5 & 5 & 5 & 5 & 5 & 5 \\
\hline
\multirow{2}{*}{ST finetune} & BLEU & 27.91 & 32.12 & 39.04 & 27.69 & 32.39 & 33.96 & 26.30 & 19.02 \\
& $\alpha$ & 4 & 4 & 5 & 4 & 4 & 4 & 4 & 3 \\
\hline
\end{tabular}
\caption{The choice for hyperparameters and the corresponding MT \& ST performance in the training steps of SimRegCR without external MT datasets.}
\label{tab:12}
\end{table*}

\begin{table*}[h]
\centering
\begin{tabular}{c | c | c c c c c c c c} 
Training Stage & & \texttt{de} & \texttt{es} & \texttt{fr} & \texttt{it} & \texttt{nl} & \texttt{pt} & \texttt{ro} & \texttt{ru} \\
\hline
\hline
MT pretrain$^{\dagger}$ & Baseline & 29.61 & 31.98 & 40.59 & 26.30 & 30.58 & 31.83 & 23.48 & 18.65 \\
\hline
MT finetune & Baseline & 33.59 & 37.78 & 45.93 & 32.74 & 37.06 & 38.81 & 29.05 & 22.11 \\
\hline
\multirow{3}{*}{ST finetune} & Baseline & 27.33 & 31.70 & 38.04 & 26.29 & 29.77 & 31.73 & 23.43 & 18.23 \\
\cline{2-10}
& BLEU & 28.96 & 33.04 & 39.37 & 27.30 & 32.22 & 33.51 & 26.00 & 19.41 \\
& $\alpha$ & 3 & 3 & 2 & 3 & 3 & 4 & 4 & 3 \\
\hline
\end{tabular}
\caption{The choice for hyperparameters and the corresponding MT \& ST performance in the training steps of SimRegCR$^-$ with external MT datasets. $\dagger$ denotes the training procedure is performed on the external MT dataset.}
\label{tab:13}
\end{table*}

\begin{table*}[h]
\centering
\begin{tabular}{c | c | c c c c c c c c} 
Training Stage & & \texttt{de} & \texttt{es} & \texttt{fr} & \texttt{it} & \texttt{nl} & \texttt{pt} & \texttt{ro} & \texttt{ru} \\
\hline
\hline
\multirow{2}{*}{MT pretrain$^{\dagger}$} & BLEU & 30.02 & 32.10 & 40.62 & 28.24 & 33.08 & 34.02 & 24.99 & 19.28 \\
& $\alpha$ & 0.5 & 0.25 & 0.125 & 3 & 3 & 2 & 2 & 0.5 \\
\hline
\multirow{2}{*}{MT finetune} & BLEU & 34.11 & 37.97 & 46.95 & 33.86 & 38.67 & 40.09 & 32.23 & 22.45 \\
& $\alpha$ & 1 & 0.25 & 3 & 5 & 5 & 3 & 3 & 3 \\
\hline
\multirow{2}{*}{ST finetune} & BLEU & 29.23 & 32.97 & 39.98 & 28.16 & 32.68 & 34.24 & 26.66 & 20.09 \\
& $\alpha$ & 3 & 3 & 3 & 3 & 3 & 4 & 3 & 4 \\
\end{tabular}
\caption{The choice for hyperparameters and the corresponding MT \& ST performance in the training steps of SimRegCR with external MT datasets. $\dagger$ denotes the training procedure is performed on the external MT dataset.}
\label{tab:14}
\end{table*}

\begin{table*}[h]
\centering
\begin{tabular}{c | c | c c c c} 
Training Stage & & \texttt{de} & \texttt{es} & \texttt{fr} & \texttt{ru} \\
\hline
\hline
MT pretrain$^{\dagger}$ & Baseline & 29.37 & 32.91 & 41.33 & 18.07 \\
\hline
MT finetune & Baseline & 33.78 & 37.53 & 45.99 & 21.67 \\
\hline
\multirow{3}{*}{ASR \& MT finetune} & Baseline & 0.47 & 0.43 & 0.43 & 0.07 \\
\cline{2-6}
& BLEU & 25.10 & 26.99 & 34.59 & 15.56 \\
& $\beta$ & 30 & 45 & 20 & 35 \\
\hline
\end{tabular}
\caption{The choice for hyperparameters and the corresponding MT \& ST performance in the training steps of SimZeroCR with external MT datasets. $\dagger$ denotes the training procedure is performed on the external MT dataset.}
\label{tab:15}
\end{table*}

\end{document}